%% file: main.tex
\documentclass[letterpaper, 10 pt, conference]{ieeeconf}  

\usepackage{xcolor}
\usepackage{graphicx}
\usepackage{caption}
\usepackage{subcaption}
\usepackage{amsmath}
\usepackage{hyperref}
\usepackage{balance}
\usepackage{amsfonts}

\usepackage[ruled,linesnumbered]{algorithm2e}
\usepackage{algorithmicx}
\usepackage{booktabs}
\makeatletter
\let\NAT@parse\undefined
\makeatother
\usepackage{verbatim}

\algnewcommand{\LeftComment}[1]{\Statex \(\triangleright\) #1}

\IEEEoverridecommandlockouts                              

\title{\LARGE \bf
Model-Predictive Control of Blood Suction for Surgical Hemostasis using Differentiable Fluid Simulations}

\author{Jingbin Huang$^{\dagger,1}$ \IEEEmembership{Student Member, IEEE}, Fei Liu$^{\dagger,1}$, Florian Richter$^{1}$ \IEEEmembership{Student Member, IEEE}\\ and Michael C. Yip$^1$ \IEEEmembership{Member, IEEE} 
\thanks{$\dagger$Equal contributions}
\thanks{$^1$Jingbin Huang, Fei Liu, Florian Richter, and Michael C. Yip are with the Department of Electrical and Computer Engineering, University of California San Diego, La Jolla, CA 92093 USA. {\tt\small \{jih023, f4liu, frichter, yip\}@ucsd.edu}}%
}

\begin{document}

\maketitle
\thispagestyle{empty}
\pagestyle{empty}

\begin{abstract}
\input{sections/abstract}
\end{abstract}

\section{INTRODUCTION}
\input{sections/introduction}

\section{RELATED WORKS}
\input{sections/related_works}

\section{METHODS}
\input{sections/methods}

\section{EXPERIMENTS AND RESULTS}
\input{sections/experiments}


\section{DISCUSSION AND CONCLUSION}
\input{sections/discussion_and_conclusion}

\balance
\bibliographystyle{ieeetr}
\bibliography{references}

\end{document}

%% file: sections/abstract.tex
Recent developments in surgical robotics have led to new advancements in the automation of surgical sub-tasks such as suturing, soft tissue manipulation, tissue tensioning and cutting.
However, integration of dynamics to optimize these control policies for the variety of scenes encountered in surgery remains unsolved.
Towards this effort, we investigate the integration of differentiable fluid dynamics to optimizing a suction tool's trajectory to clear the surgical field from blood as fast as possible.
The fully differentiable fluid dynamics is integrated with a novel suction model for effective model predictive control of the tool.
The differentiability of the fluid model is crucial because we utilize the gradients of the fluid states with respect to the suction tool position to optimize the trajectory.
Through a series of experiments, we demonstrate how, by incorporating fluid models, the trajectories generated by our method can perform as good as or better than handcrafted human-intuitive suction policies.
We also show that our method is adaptable and can work in different cavity conditions while using a single handcrafted strategy fails. 





%% file: sections/introduction.tex
Automating robotic surgeries has generated considerable interest in the robotics research community \cite{yipDasJournal} given the development of commercially available surgical robotic systems like the da Vinci Surgical System and open-source platforms such as the da Vinci Research Tool Kit (dVRK) \cite{kazanzides2014open}, Some important goals of surgical automation research are to develop algorithms and systems that can leverage the precision of robots to improve the safety of operations, reduce surgeon fatigue by taking over repetitive tasks, and improve access to surgical care for communities where clinician availability is lacking. Initial research has been done towards automating surgical sub-tasks such as soft tissue manipulation \cite{Shin_2019,attanasio2020autonomous}, tensioning and cutting \cite{thananjeyan2017multilateral,le2018semi}, suturing \cite{zhong2019dual, shademan2016supervised, pedram2017autonomous}, and debridement removal \cite{richter2019open, pedram2019toward}.

An important component to  automating of surgical tasks is understanding the interaction of the surgical tool and the environment.
Modelling the interaction between surgical tools and the environment has previously been done using data driven approaches in the form of reinforcement learning \cite{thananjeyan2017multilateral} and learning from demonstration \cite{murali2015learning}.
While these methods provide good initial results, the policies produced are limited by the variation in the scenarios in which the policy is trained upon, and those policies may fail in new environments. 
In order to develop more generalizable autonomous control policies, an online model-predictive approach is preferred. we consider differentiable dynamics where gradients can be computed through the predictive models, hence allowing for trajectories to be efficiently optimized on a per-scenario basis.
This allows the robot to adapt its trajectories, in real-time, to the constantly deforming and highly variable surgical environment.
Such differential dynamics models have been demonstrated in \cite{ummenhofer2019lagrangian}. \cite{li2018learning}, and \cite{schenck2018spnets}, but have not been applied to surgical robotics.



\begin{figure}
    \centering
    \vspace{1.8mm}
    \begin{subfigure}{0.47\textwidth}
        \includegraphics[width=1.0\textwidth]{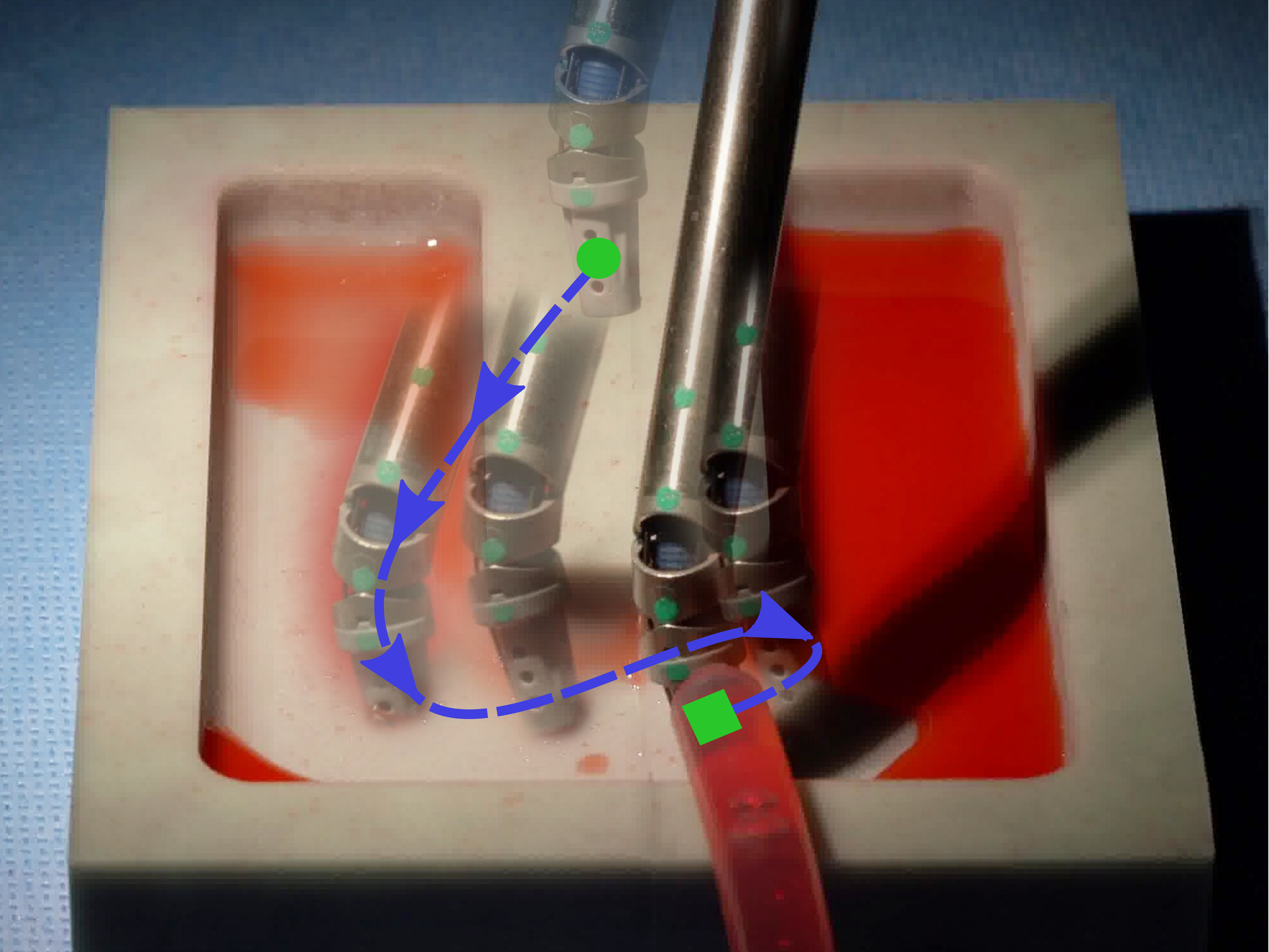}
        \label{fig:case2_tool_motion_real}
    \end{subfigure}
    
    \caption{A suction trajectory generated using our model predictive control algorithm being deployed in a silicon rubber cavity. By explicitly incorporating fluid dynamics with gradients back-propagated through a position-based fluid simulation, our method is able to efficiently optimize trajectories to rapidly clear the surgical field from blood in a model-predictive manner.
    }
    \label{fig:case2_real_motion_and_trajectory}
\end{figure}


In this work we will investigate integration of differentiable dynamics from the perspective of fluids and autonomous suctioning.
Suctioning is the first essential step to achieving hemostasis, which is the process of suctioning blood, finding the bleed, and closing a ruptured vessel during a surgery.
Ruptured vessels can occur at any moment and require an immediate reaction.
By clearing the surgical field from blood in an quick manner, a surgeon can identify the rupture vessel and close it effectively.
To accomplish this, we develop a model-predictive control (MPC) strategy utilizing differentiable blood fluid dynamics for the surgical scene.  Specifically, we present the following contributions:
\begin{enumerate}
    \item an end-to-end differentiable position-based blood fluid simulation,
    \item a differentiable model of suction that can be embedded into the fluid simulation, and
    \item a model-predictive control algorithm that incorporates blood fluid dynamics in autonomous surgical robotics.
\end{enumerate}
Using a natively differentiable fluid model provides a robust method for integrating the behavior of flowing blood into an optimization framework. We study and evaluate the proposed method through simulated experiments to highlight the advantages and generalizability of the resulting control policy for blood suction.
We also provide an implementation of the method on the da Vinci Research Kit (dVRK) \cite{kazanzides2014open} to show the real-world effectiveness of the proposed methods.


%% file: sections/related_works.tex
Fluid modeling and simulation have been studied extensively for applications in the natural sciences and computer graphics. For example, smoothed-particle hydrodynamics (SPH) is a method originally developed in 1977 for astrophysical applications \cite{gingold1977smoothed} but have now been widely applied in fluid animations \cite{egst.20141034}. Recent advancements in SPH  \cite{solenthaler2009predictive,akinci2013versatile} aim to improve the accuracy and speed of the simulation. Furthermore, SPH can be embedded into a position-based dynamics simulations to create position-based fluids (PBF), which allows for greater stability when using large timesteps \cite{macklin2013position}. However, these methods are only for the forward simulation of fluid, and do not provide gradients. 

More recently, several methods of differentiable fluid models have been developed. For example, in \cite{ummenhofer2019lagrangian}, the authors introduce a Lagrangian fluid simulation using a new type of convolutional neural network.
Deep neural networks have also been used to learn fluid dynamics in \cite{li2018learning}. 
Schenck and Fox implement PBF using novel fluid dynamic operators to interface with neural networks layers \cite{schenck2018spnets}. 
All of these methods provide naturally differentiable fluid models and are capable of optimizing downstream tasks through their gradients.
In this work, we use a similar approach to computing the gradients as \cite{schenck2018spnets} with an extension to differentiable modelling for surgical suction.
Combining the differentiable modelling with MPC allows for the initial blood clearing task of hemostatis to be solved efficiently.

%% file: sections/methods.tex
\begin{algorithm}[t]
\SetAlgoLined

$\mathbf{v}_i = \mathbf{v}_i + \Delta t f_{ext}(\mathbf{x}_i)$ \Comment{{\scriptsize apply external forces}} \\
$\mathbf{x}_i^* = \mathbf{x}_i + \Delta t \mathbf{v}_i$ \Comment{{\scriptsize predict position}} \\
$\mathbf{n}_i \leftarrow \mathrm{FindNeighboringParticles}(\mathbf{x}_i^*)$  \\

\While{iter $<$ SolverIterations}{
$\lambda_i \leftarrow \mathrm{computeDensityConstraintMultiplier}(\mathbf{n}_i)$  \\
$\Delta \mathbf{x}_i\leftarrow \mathrm{solveDensityConstraint}(\lambda_i)$  \\
$\Delta_{u,i}, \Delta_{p,i} \leftarrow \mathrm{solveSuctionDisplacements}(\mathbf{x}_e)$ \\
$\mathbf{x}_i^* = \mathbf{x}_i^* + \Delta \mathbf{x}_i + \mathbf{\hat{y}} \Delta_{u,i} + \Delta_{p,i}$ \\
$\mathbf{x}_i^* \leftarrow \mathrm{computeBoundaryCollision}(\mathbf{x}_i^*)$  \\
}
$\mathbf{x}_i =  \mathbf{x}_i^* $  \Comment{{\scriptsize update position}} \\
$\mathbf{v}_i = \left( \mathbf{x}_i^* - \mathbf{x}_i \right)/\Delta t$ \Comment{{\scriptsize update velocity}} \\
\caption{PBF simulation loop for particles and suction nozzle at positions $\mathbf{x}_i$ and $\mathbf{x}_e$ respectively}
\label{alg:basic_PBF}
\end{algorithm}

At a high level, our method of incorporating fluid dynamics into autonomous suction control involves three main components: (a) a differentiable fluid model, (b) a suction model between fluid and a suction tool, and (c) a model-predictive controller. The key insights to making the overall autonomous suction work well is that, first, we make the fluid model differentiable by viewing its operations as computational graphs and using back-propagation; then, we make the suction model continuous and differentiable force field that removes particles from the simulation. This allows the entire fluid and suction modeling to be incorporated with the robot manipulator model as a complete, differentiable system of equations that a MPC scheme can effectively and efficiently solve.

\subsection{Differentiable position-based fluid simulation}






    
    
    


Position-based fluids (PBF) is used as the basis of the fluid simulation because it has good stability even when using large timesteps \cite{macklin2013position}. In the PBF problem, we consider fluids to be  represented by a set of $N$ particles in 3D Cartesian space, $\mathbf{X}\in \mathbb{R}^{3\times N}$. Each particle has position $\mathbf{x}_i \in \mathbb{R}^3$ and velocity $\mathbf{v}_i \in \mathbb{R}^3$. To model fluid physics, these particles are given position-based constraints. In PBF, a density constraint acts on the particle and its neighbors such that they maintain a proximity with one another that as-closely-as-possible matches the resting density of the fluid.
The PBF simulation loop is outlined in Algorithm \ref{alg:basic_PBF}.


In order to make this simulation differentiable, all mathematical operations applied to these particles are treated as computation graphs.
From the graph, a gradient back-propagation is used between each operation of the solver. Using this technique, we greatly simplify the application of the chain-rule for complicated operations because the gradient of each step can be computed independently using cached intermediate results. 

\begin{figure}[t]
    \centering
    \vspace{1.8mm}
    \includegraphics[trim=0cm 1cm 0cm 0.5cm, clip, width=0.47\textwidth]{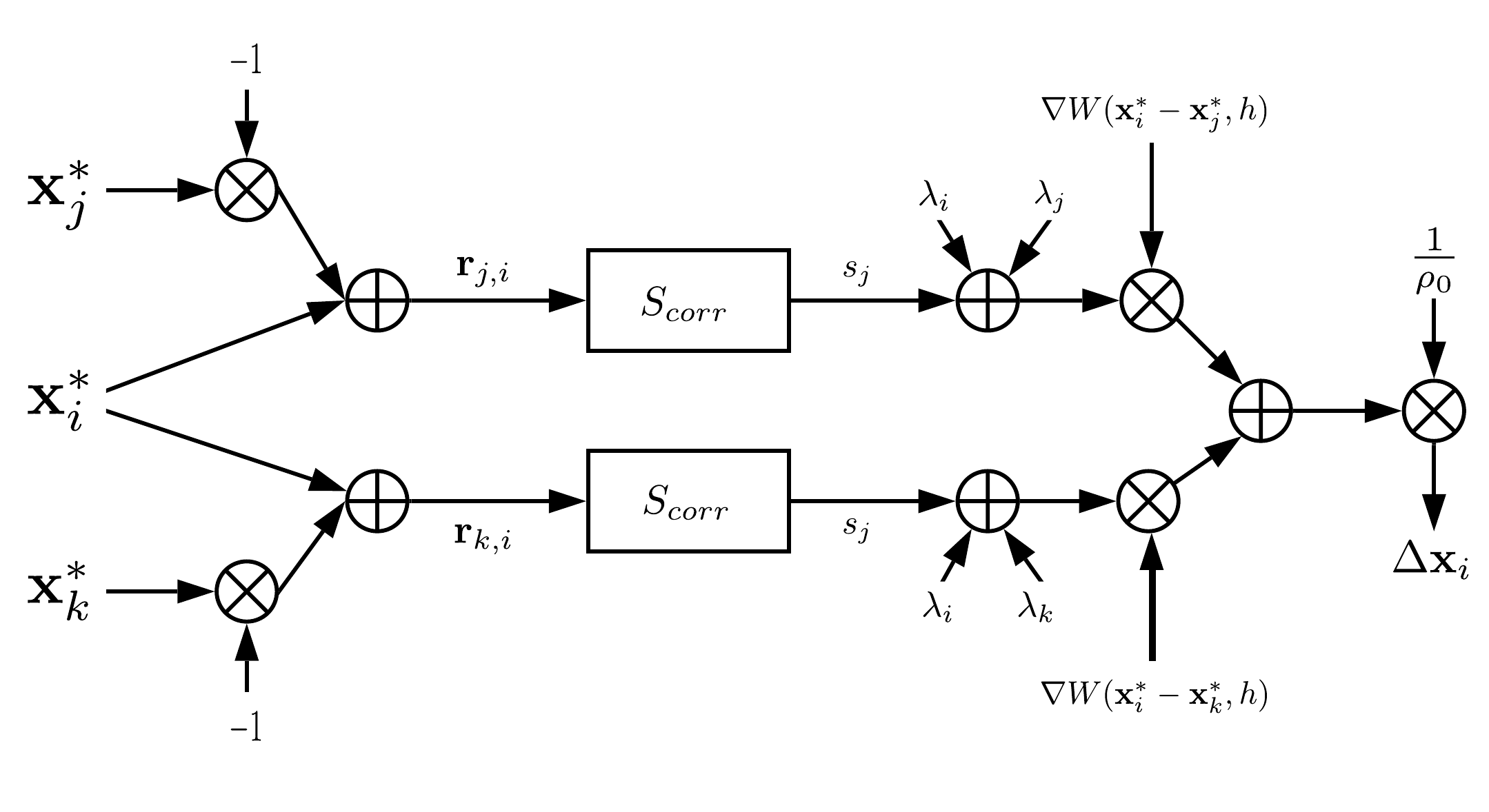}
    \caption{An example computation graph corresponding to (\ref{eq:fluid_density_constraint}). The graph represents the operations for computing the position correction, $ \Delta \mathbf{x}_i $, to satisfy the fluid density constraints.}
    \label{fig:compute_position_correction_graph}
\end{figure}

As an example, consider the series of operations in Step 6 of Algorithm \ref{alg:basic_PBF} which computes particle position offsets to satisfy the fluid density constraint.
The formula for computing the correction is given in \cite{macklin2013position} as
\begin{equation}
    \Delta \mathbf{x}_i = \frac{1}{\rho_0} \sum_j (\lambda
    _i + \lambda_j + s_{corr}) \nabla W(\mathbf{x}_i - \mathbf{x}_j, h)
    \label{eq:fluid_density_constraint}
\end{equation}
where $\rho_0$ is the rest density of the fluid, $\lambda$ is the density constraint multiplier for each particle $i = 1...M$, $s_{corr}$ is an artificial pressure term used to maintain surface tension, $\nabla W$ is the gradient of the Spiky kernel, given as 
\begin{equation}
    \nabla W (\mathbf{r}, h) = \frac{45}{\pi h^6} (h-||\mathbf{r}||_2)^2 \frac{\mathbf{r}}{||\mathbf{r}||_2}
\end{equation}
and $h$ is the neighborhood radius, which is a parameter of the simulation.
The computational graph representing these operations is shown in Figure \ref{fig:compute_position_correction_graph}. Because the values of $\lambda$ are also functions of the particle positions from Step 5 in Algorithm \ref{alg:basic_PBF}, the total gradient of $\Delta \mathbf{x}_i$ has contributions from multiple paths: through $S_{corr}$, through $\lambda$, and through $\nabla W$. With the back-propagation view, gradients through each of these paths can be computed independently using cached intermediate values and subsequently combined to obtain the end-to-end gradient.



\subsection{Suction model}


The suction force from the surgical tool is modelled as a continuous and differentiable force field that pulls nearby particles towards the nozzle of the tool and adds large vertical displacement to the particles within some effective suction area. Note that we define $y$ as the vertical direction while the $x \text{-} z$ plane represents the horizontal plane. 
%
The upward displacement field is in the vertical direction, as this is the typical orientation of a suction tool attacking pooling blood. It is modelled after a 2D-Gaussian probability density function (PDF) over the $x\text{-}z$ plane due to its differentiable properties. The magnitude of the upward displacement field experienced by particle $i$ is proportional to the value of the PDF at $([\mathbf{x}_{i}]_x,[\mathbf{x}_{i}]_z)$. 
For a particle $i$ with position $\mathbf{x}_i$, and a suction nozzle at position $\mathbf{x}_e \in \mathbb{R}^3$, the upward displacement, $\Delta_u$, is computed as:
\begin{equation}
\Delta_{u,i} = \mathcal{K} \cdot \frac{\text{exp} \left( -\frac{\left([\mathbf{x}_{i}]_x-[\mathbf{x}_{e}]_x \right)^2}{2\sigma_x^{2}}   - \frac{\left([\mathbf{x}_{i}]_z-[\mathbf{x}_{e}]_z \right)^2}{2\sigma_z^{2}}   \right) }{\sqrt{(2 \pi)^2 \sigma_x^2 \sigma_z^2}}
\end{equation}
where $[\cdot]_x$, $[ \cdot]_y$, and $[ \cdot ]_z$ are the $x,y,z$ coordinates of the position vectors and $\sigma_x$ and $\sigma_z$ are the standard deviations of the 2D Gaussian PDF. The standard deviation controls how narrow the suction region is. Finally, $\mathcal{K} > 0$ controls the size of the upward displacement and is adjusted to control the strength of the suction.
For the field that pulls particles towards the end-effector at position $\mathbf{x}_e$, the $i$-th particle will experience a displacement given by
\begin{equation}
    \Delta_{p,i} = \frac{\mathbf{x}_{e} - \mathbf{x}_{i}}{||\mathbf{x}_{e} - \mathbf{x}_{i}||_2} \cdot \frac{1}{||\mathbf{x}_{e} - \mathbf{x}_{i}||_2^2 + d}
\end{equation}
where $d$ is a constant that limits the maximum displacement when the particle is very close to the center of the end-effector as well as to avoid division by 0. 
Finally, the upward field and the field that pulls particles towards the end-effector are summed and added to the particle positions along with the PBF solver corrections as shown in line 7 of Algorithm \ref{alg:basic_PBF}, where $\mathbf{\hat{y}}$ is a normalized vector that describes the direction of upward displacement in the simulation.
This formulation of suction, embedded in a completely differentiable fluid model, enables efficient and stable calculation of gradients from the particle states to the suction tool position. 

\begin{algorithm}[t]
\SetAlgoLined

$\mathbf{X}_{t} \leftarrow \{\mathbf{x}_{1,t}, ~ \mathbf{x}_{2,t}, \cdots\}$ \Comment{{\scriptsize get particle positions}} \\
$\mathbf{V}_{t} \leftarrow \{\mathbf{v}_{1,t}, ~ \mathbf{v}_{2,t}, \cdots\}$ \Comment{{\scriptsize get particle velocities}} \\
$\mathcal{U} = \{\mathbf{x}_{e,t}, ..., \mathbf{x}_{e,h}\} \leftarrow \mathbf{x}_{e,t+h-1}^*$ \Comment{{\scriptsize Initialize trajectory using previous end-effector position}} \\

\For{numIterations}{
$\mathrm{initializePBFSimulation}(\mathbf{X}_{t}, \mathbf{V}_{t})$  \Comment{{\scriptsize Setup environment}} \\
$\mathbf{X}_{t+h}, \mathbf{V}_{t+h}\leftarrow \mathrm{runPBFSimulation}(h)$  \Comment{{\scriptsize Run forward PBF for $h$ time-steps}} \\
$ L = \mathrm{computeLoss}(\mathbf{X}_{t+h})$  \Comment{{\scriptsize According to Eq. \ref{eq:loss}}} \\
$ \mathcal{G} = \frac{\partial L}{\partial \mathcal{U}}$  \Comment{{\scriptsize Gradient  back-propagation }} \\
$ \mathcal{U} = \mathcal{U} - \alpha \mathcal{G} $  \Comment{{\scriptsize Gradient descent}}
}


$\mathbf{x}_{e,t}^* =  \mathbf{x}_{e,t} $  \Comment{{\scriptsize Optimal MPC control}} \\

\caption{Model predictive end-effector control
}
\label{alg:MPC}
\end{algorithm}

\subsection{Model predictive control formulation}


The overall goal of the proposed control problem is to achieve hemostasis as quickly as possible, which corresponds to when all particles in the modelled scene have been lifted from the underlying tissue surfaces of the scene.
This will be formatted as an optimization problem where MPC \cite{camacho2013model} is used to generate a trajectory that the suction nozzle should follow to optimally clear the surgical field from fluids.
Let the control input to the MPC be denoted as $\mathcal{U}=\{\mathbf{x}_{e,t}\}$ where $\mathbf{x}_{e,t}$ is the position of the suction nozzle at time $t$.
The output of the system is the set of particles $\mathbf{X}_t = \{\mathbf{x}_{i,t}\}$.
The MPC is computed over a short horizon, $h$, to find a control trajectory $\{\mathbf{x}_{e,t}^*, ..., \mathbf{x}_{e, t+h}^*\}$ which minimizes the loss over the time frame $[t,t+h]$.
The optimization problem finds the set of control inputs $\mathcal{U}$ that minimizes the loss
\begin{equation}
\label{eq:loss}
\begin{split}
    \min_{\mathcal{U}}   L = \sum_t \sum_i l(\mathbf{x}_{i,t})
\end{split}
\end{equation}
over all timesteps and particles.
The cost function is set to
\begin{equation}
    l(\mathbf{x}_{i,t})= 
    \begin{cases}
        \frac{1}{2} \| y_{goal} - \left[ \mathbf{x}_{i,t} \right]_y \|_2^2, \quad & \text{if $\left[ \mathbf{x}_{i,t} \right]_y < y_{goal}$} \\
        0, \quad & \text{otherwise}
    \end{cases}
\end{equation}
where $y_{goal}$ is a target height set above the surgical cavity. Suction completion is defined as such so that it is continuously differentiable. In MPC, only $\mathbf{x}_{e,t}^*$ is applied to the system, and then another entire horizon length of control is computed again starting at $t+1$.
This routine is detailed in Algorithm \ref{alg:MPC}.



To determine a good initial suction point, $\mathbf{x}_{e,0}^*$, a Monte-Carlo approach is utilized.
The process involves sampling many possible starting points from the fluid particle positions, performing roll-outs using the MPC algorithm with these samples over a look-ahead window $m$, and finally selecting the best performing one with regards to removing the most fluid.
An outline is of this procedure is shown in Algorithm \ref{alg:initial}.
\begin{algorithm}[t]
\SetAlgoLined

$\mathbf{X}_{0} \leftarrow \{\mathbf{x}_{1,0}, ~ \mathbf{x}_{2,0}, \cdots\}$ \Comment{ {\scriptsize get particle positions} }\\
$\mathbf{V}_{0} \leftarrow \{\mathbf{v}_{1,0}, ~ \mathbf{v}_{2,0}, \cdots\}$ \Comment{ {\scriptsize get particle velocities} }\\
$\mathbf{S} = \{\mathbf{s}_1, \mathbf{s}_2, \cdots \mathbf{s}_{N}\} \leftarrow \mathbf{X}_{0} $ \Comment{{\scriptsize Uniformly sample $N$ point from $\mathbf{X}_{0}$}} \\

\For{$\mathbf{s}_k$ in $\mathbf{S}$}{
$\mathbf{X}_{t+m}, \mathbf{V}_{t+m}\leftarrow \mathrm{runAlgorithm2}(\mathbf{s}_k, m)$  \Comment{ {\scriptsize Run MPC with $\mathbf{S}$ as initial end-effector position for $m$ time-steps} } \\
$ \mathbf{r}_{k} \leftarrow \lbrace \mathbf{x}_{j,t+m} | \left[ \mathbf{x}_{i,t+m} \right]_y < y_{goal}, \mathbf{x}_{i,t+m} \in \mathbf{X}_{t+m} \rbrace $
\Comment{ {\scriptsize The remaining set of particles not reaching target} } \\
$n_k = \mathrm{card}{(\mathbf{r}_{k})}$
\Comment{ {\scriptsize Number of particles in $\mathbf{r}_{k}$} }
}

$ \mathbf{s}_* \leftarrow \min_{\mathbf{s}_k \in \mathbf{S}}  n_*$  \Comment{Optimal initial suction point} \\
\caption{Selection of initial suction point
}
\label{alg:initial}
\end{algorithm}








%% file: sections/experiments.tex
The efficiency of the proposed suction control algorithm is tested in both simulated scenes and a real world example.
Performance is evaluated by measuring how quickly the fluid is removed.
It is also compared against hand-crafted trajectories to highlight the generalizability of the proposed methods to different scenarios.

\begin{figure}[t]
    \centering
    \begin{subfigure}{0.15\textwidth}
        \includegraphics[width=1.0\textwidth]{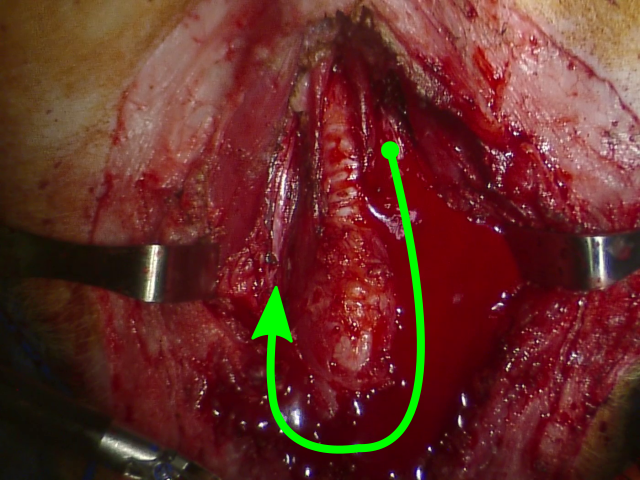}
        \label{fig:cavity1_real}
    \end{subfigure}
    \begin{subfigure}{0.15\textwidth}
        \centering
        \includegraphics[width=1.0\textwidth]{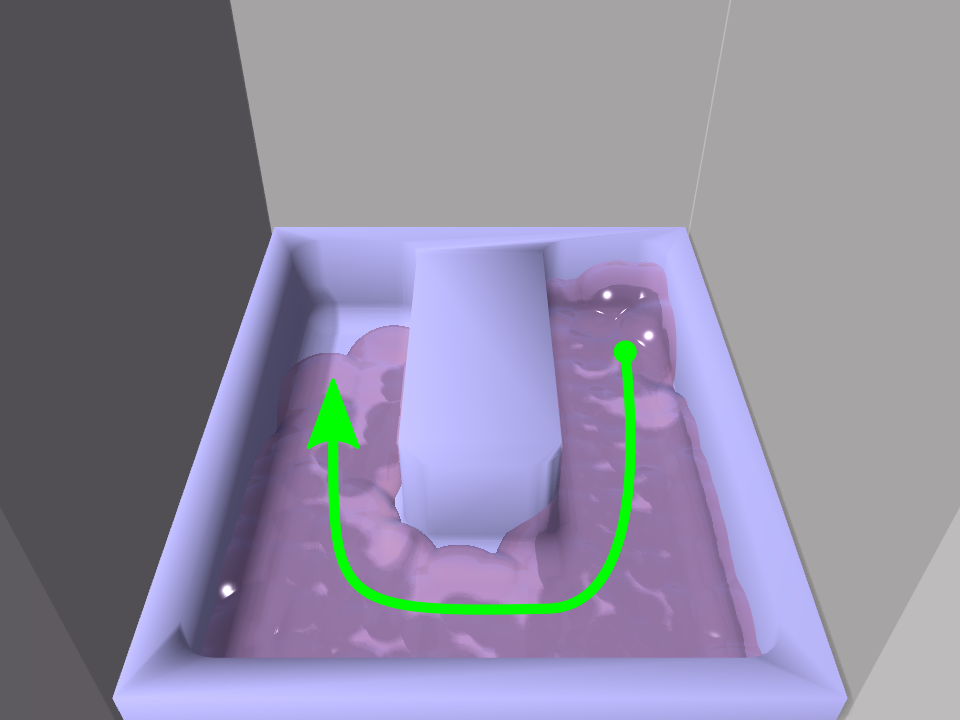}
        \label{fig:cavity1_sim}
    \end{subfigure}
    \begin{subfigure}{0.15\textwidth}
        \centering
        \includegraphics[width=1.0\textwidth]{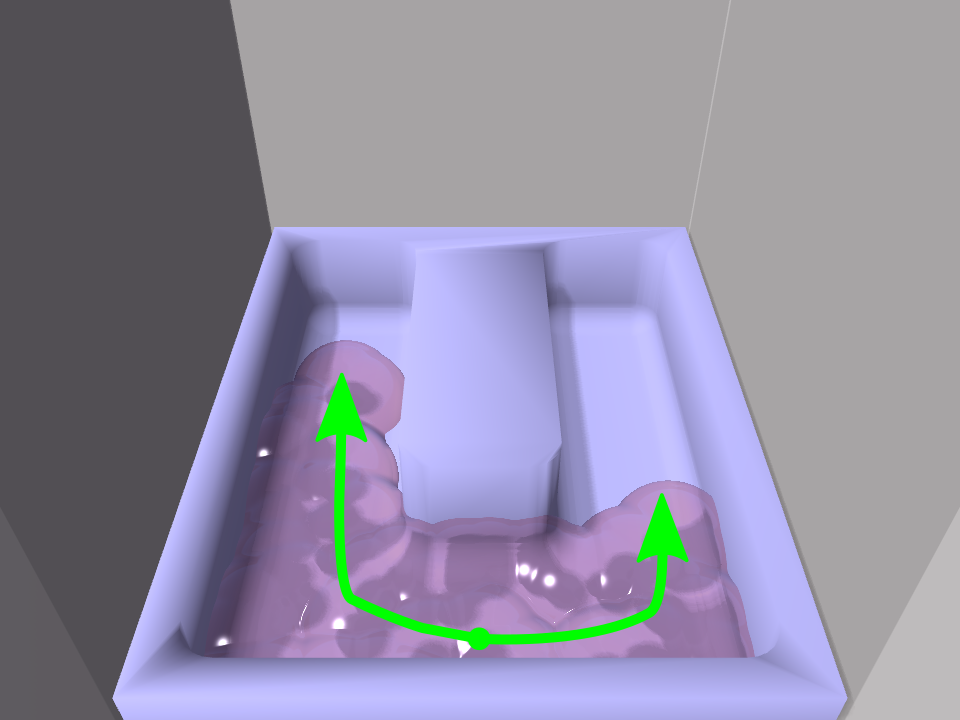}
        \label{fig:cavity3_sim}
    \end{subfigure}
    \caption{Comparison between a real surgical cavity and our simulated cavity. \textbf{Left:} the scene of a thyroidectomy conducted on a pig, where a rupture occurred on the carotid artery. \textbf{Middle:} a simulated cavity that has the same general flow pattern as the real scene, denoted as \textit{case 1}. \textbf{Right:} same cavity but with a different blood emission point, denoted as \textit{case 2}. The green paths and arrows denote the general direction of blood flow. }
    \label{fig:cavity_side_by_side}
\end{figure}

\begin{figure}[t]
    \centering
    \vspace{1.8mm}
    \begin{subfigure}{0.235\textwidth}
        \centering
        \includegraphics[trim=0cm 0cm 0cm 0cm, clip,width=1.0\textwidth]{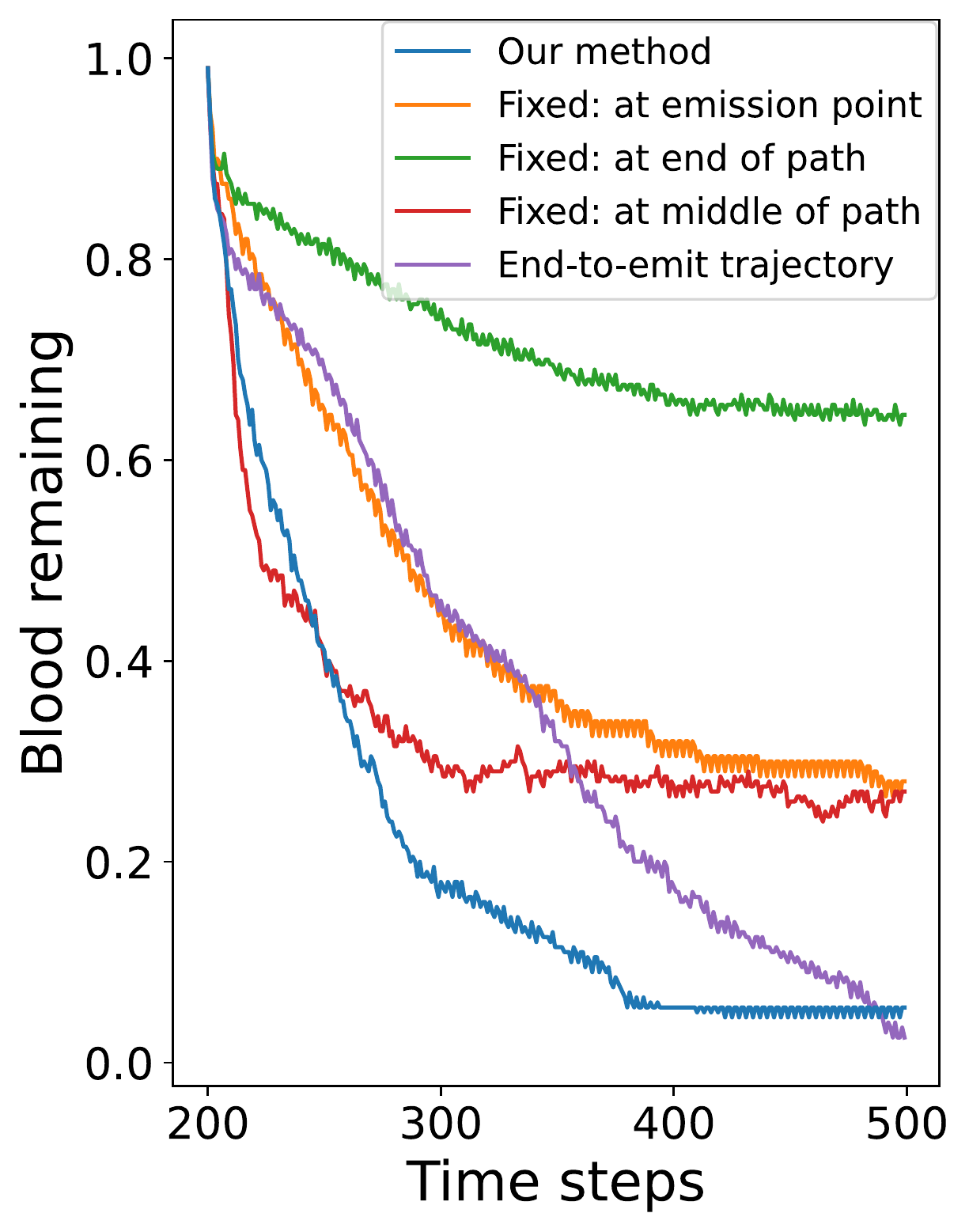}
        \label{fig:sim_suction_curve_case1}
    \end{subfigure}
    \begin{subfigure}{0.235\textwidth}
        \centering
        \includegraphics[width=1.0\textwidth]{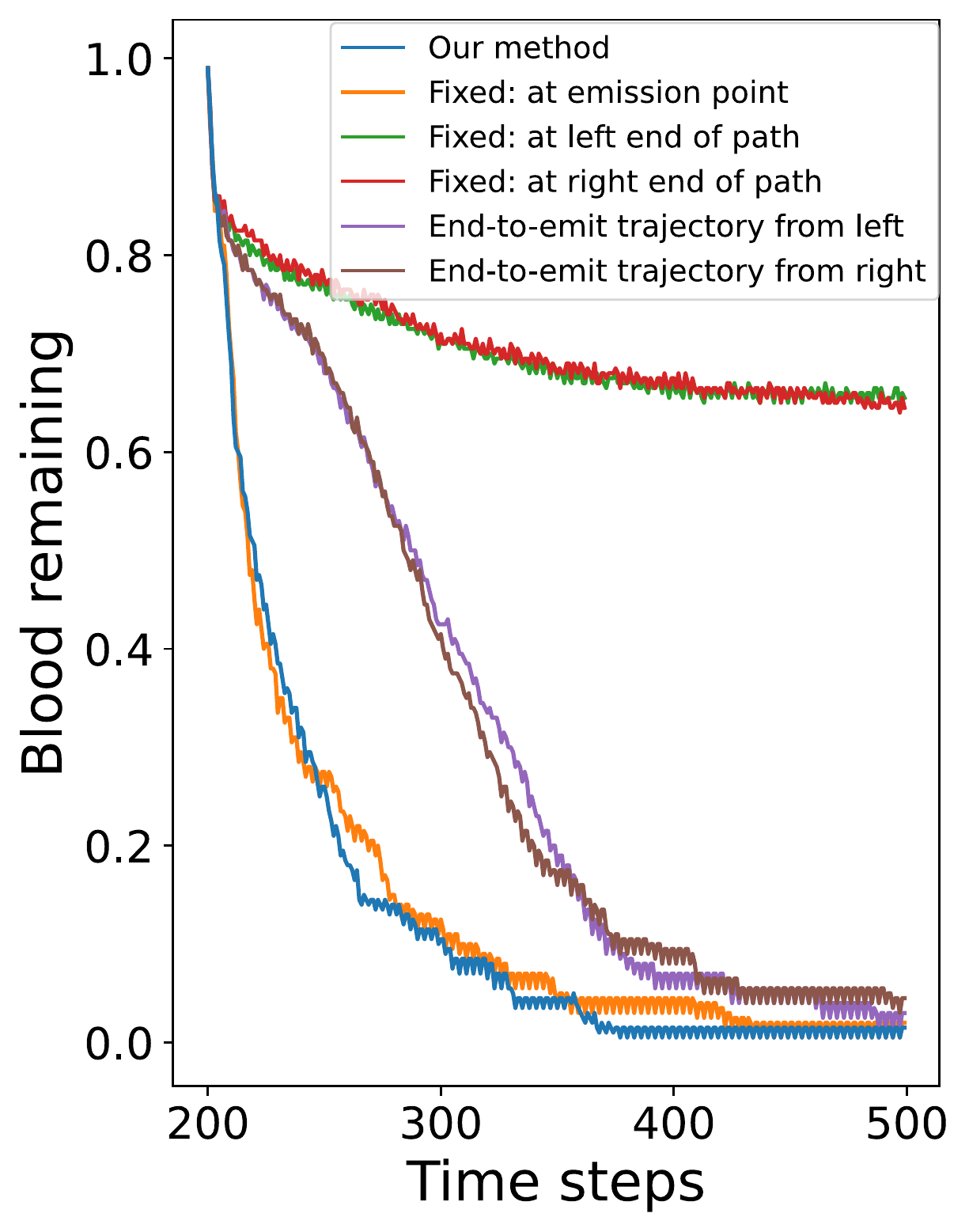}
        \label{fig:sim_suction_curve_case2}
    \end{subfigure}
    \caption{Suction curves for the simulated scenes normalized by initial volume. The time axis starts at 200 because of the warm-up period for filling the cavity. \textbf{Left:} In \textit{case 1}, both our method and the end-to-emit trajectory can reach an equilibrium where the cavity is mostly clear, with out method being much faster. \textbf{Right:} In \textit{case 2}, the fastest method are our method and suctioning at the emission point.}
    \label{fig:sim_suction_curves}
\end{figure}

\subsection{Simulated suction experiments}

Two simulated scenes used to evaluate the proposed control strategy are shown in Fig. \ref{fig:cavity_side_by_side}. These scenes are motivated by real surgical cases where a surgeon error in a live thyroidectomy caused bleeding to occur. A simulation was built to mimic this scenario, with different blood emission locations. The simulation used a time step of $\Delta_t = 0.01$s, the maximum number of particles was limited to $N=2000$, and 200 simulation steps were taken before the robot can take its first action.
The MPC controller used a horizon $h=10$, and the initial suction point selection uses a look-ahead window of $m=100$.
A total of 10 samples were used for initial point selection.
The target height was set to $y_{goal} = 10$cm, suction strength $\mathcal{K} = 100 $, $\sigma_x = \sigma_z = \sqrt{0.5}$.
The maximum change in end-effector position was also limited to 0.5mm every step to be realistic.
Finally, the gradients used a learning rate of 0.1 with normalized gradients such that their components have a maximum magnitude of 1.

Four hand-crafted control policies for controlling the suction nozzle were also developed to compare against the proposed MPC method.
They are: 1) \textit{Fixed emission}: stays at the emission point 2) \textit{Fixed end}: stays at the end point where the fluid could flow too 3) \textit{Fixed middle}: stays at the middle of the fluid flow 4) \textit{End-to-emit}: moves from the end to emission point at a constant rate.
To evaluate the performance of the methods, curves were generated by plotting the amount of remaining blood (normalized by the initial volume) against time (in simulation timesteps).
Good suction performance should have little blood at the end of the simulation.
Another metric was the time it takes to reduce the amount of blood in the cavity by 50\% and 90\%, computed as
\begin{equation}
\begin{aligned}
        \tau_{50\%} = \; & \text{min} & & t-t_0 \\
                      & \text{s.t.} & & f(t) <= 50\%
\end{aligned}
\label{eq:convergence_time}
\end{equation}
where $f(t)$ is the suction curve as a function of time, $t_0$ is the time at which suction begins, and $t_f$ is the time at which the simulation ends. The time for 90\% reduction was computed similarly. Note that the percent reduction time only makes sense if the suction policy actually reduced the amount of blood by the targeted amount. Otherwise the percent reduction time was not computed.

\begin{table}[t]
    \centering
    \vspace{1.8mm}
    \begin{tabular}{cccccc}
         \hline
         & Our & Fixed:  & Fixed:  & Fixed:  & End-to-emit\\
         & method & emission & end & middle & trajectory\\
         \hline
         Residual & 5.5\% & 28.0\% & 64.5\% & 27.0\% & \textbf{2.5}\% \\
         $\tau_{50\%}$ & \textbf{36} & 87 & -- & 23 & 90 \\
         $\tau_{90\%}$ & \textbf{162} & -- & -- & -- & 255 \\
         \hline
    \end{tabular}
    \caption{Residual of blood after trajectory is executed and percent reduction time for \textit{case 1}.
    }
    \vspace{-3mm}
    \label{tab:results_case1}
\end{table}

\begin{table}[t]
    \centering
    \begin{tabular}{ccccc}
         \hline
         & Our & Fixed:  & Fixed:  & End-to-emit \\
         & method & emission & end (avg) & trajectory (avg)\\
         \hline
         Residual & \textbf{1.5}\% & 2.0\% & 65.0\% & 3.8\% \\
         $\tau_{50\%}$ & 21 & \textbf{18} & -- & 87.5 \\
         $\tau_{90\%}$ & \textbf{102} & \textbf{102} & -- & 172.5 \\
         \hline
    \end{tabular}
    \caption{Residual of blood after trajectory is executed and percent reduction time for \textit{case 2}.
    }
    \label{tab:results_case2}
\end{table}

The results from the experiments are shown in Fig. \ref{fig:sim_suction_curves} and Table \ref{tab:results_case1} and \ref{tab:results_case2}.
As is evident by the results, the proposed method performed as good as, if not better than, the best hand-crafted policies.
In \textit{case 1}, the blood flowed around the obstacle from right to left, hence making the best hand-crafted policy, end-to-emit trajectory, slow but complete in suctioning the blood.
However, in \textit{case 2} this end-to-emit strategy failed since the blood emits closer to the middle of the cavity hence diverting in two directions.
Meanwhile the opposite behavior occurred for the fixed emission control, where it failed in \textit{case 1} but was the best strategy in \textit{case 2}. This means the hand-crafted policies had limited success only in a single scenario the proposed method successfully generalized to both cases.

\begin{figure}[t]
    \centering
    \vspace{1.8mm}
    \begin{subfigure}{0.205\textwidth}
        \centering
        \includegraphics[width=1.0\textwidth]{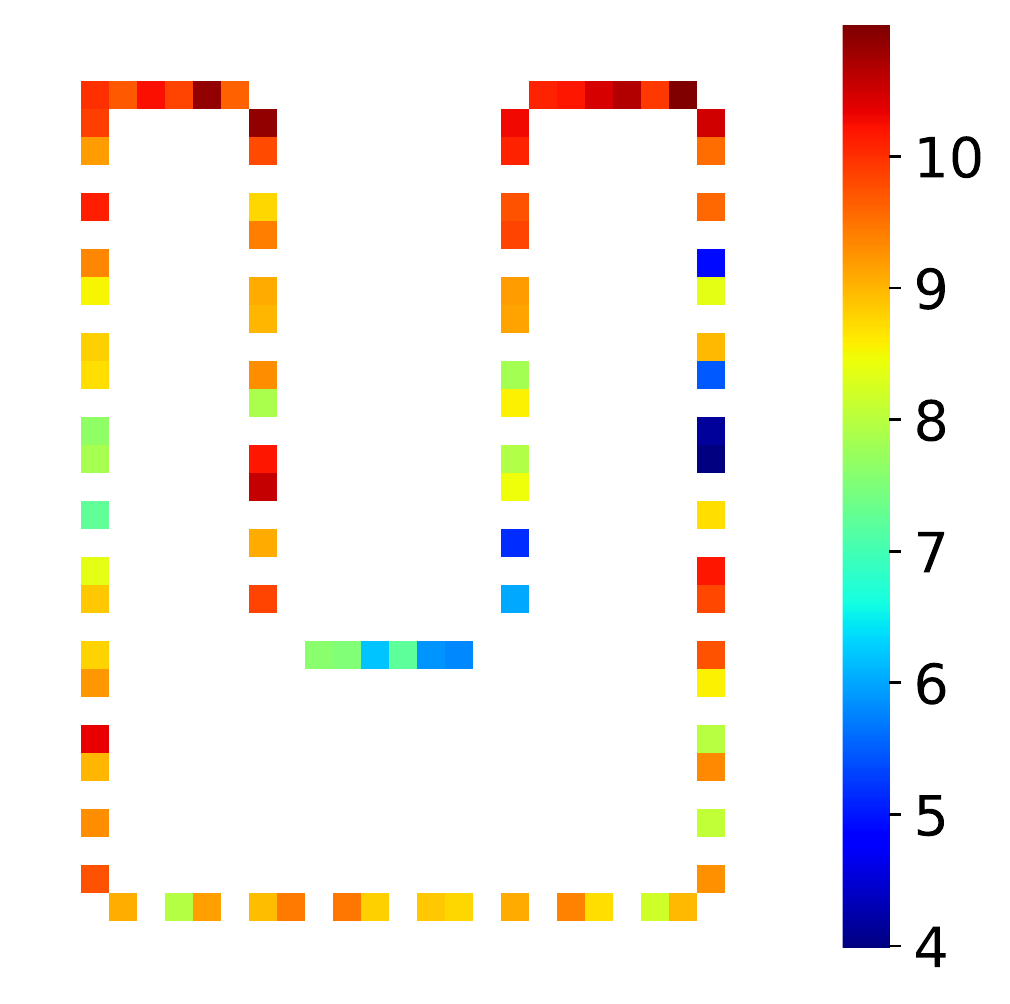}
    \end{subfigure}%
    \begin{subfigure}{0.226\textwidth}
        \centering
        \includegraphics[width=1.0\textwidth]{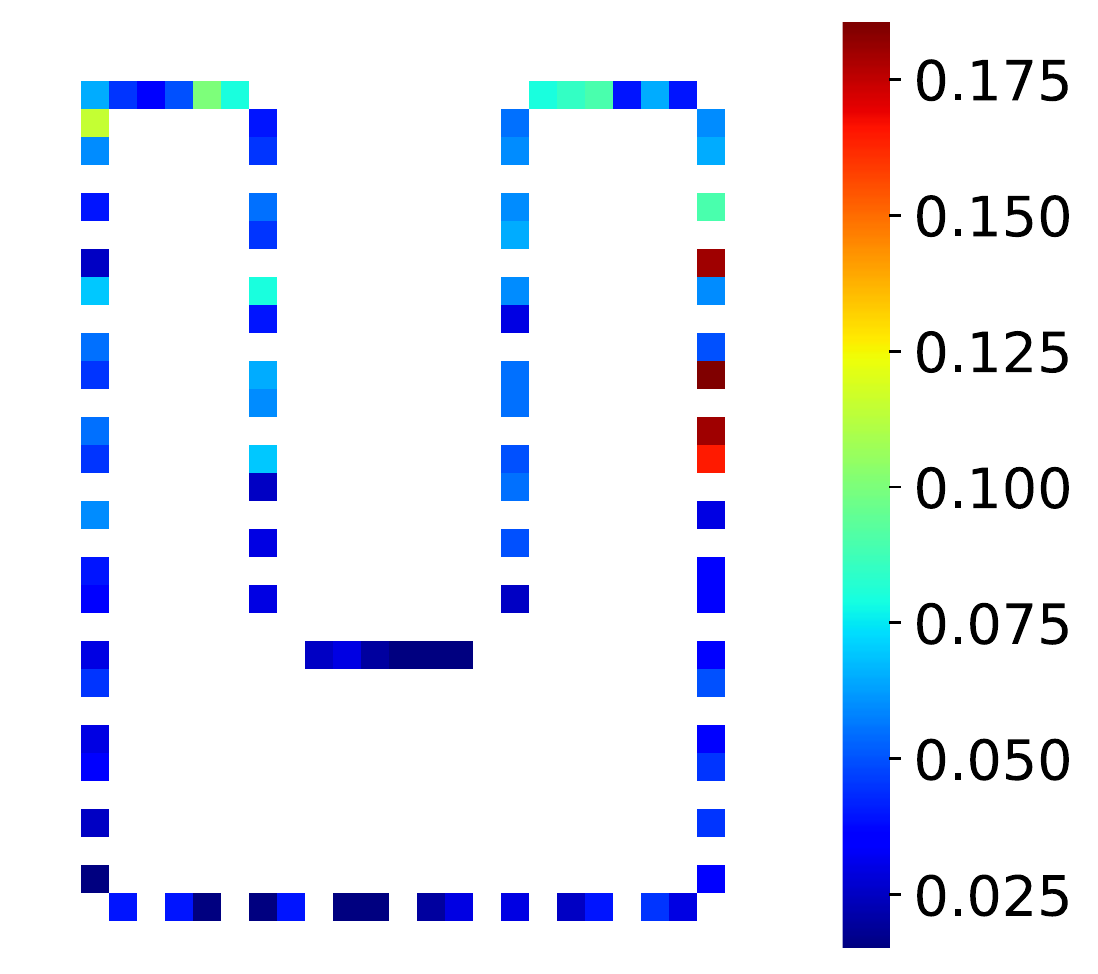}
    \end{subfigure}
    \caption{Applying our method to different emission points in the cavity. The locations of the tiles represent the locations of the emission points, while the colors denote trajectory length in $cm$ (left) and normalized amount of residual blood (right). While the trajectories vary in length as the emission point changes, they generally achieve low residuals. This demonstrates how MPC is able to adapt to different emission scenarios.
    }
    \label{fig:heatmap}
    \vspace{-1.8mm}
\end{figure}

To better understand the adaptability of the proposed method to different emission scenarios, it was tested on a large variety of emission points in the same cavity.
The emission points were evenly spaced along the sides of the cavity, with emission directions normal to the walls, hence pointing towards the inside of the cavity.
The lengths of the resulting trajectories as well as suction speed measured by 60\% convergence time are plotted as heatmaps in Fig. \ref{fig:heatmap}.
Typically the trajectories for emission points near the middle of the cavity will be shorter as the suction tool will stay near emission.
Meanwhile for emission points near the top left or right corners, the trajectory will follow more similarly to a long, end-to-emit-type path.


\subsection{Trajectory validation}


In order to validate the suction trajectories generated using our algorithm, we repeated the simulation experiments in a cavity made out of silicone rubber. 
Water with red dye was used to emulate blood flow, which was manually injected into the cavity with a syringe similar to the previously described \textit{case 1} and \textit{case 2} simulated scenes.
A Patient Side Manipulator (PSM) from the dVRK \cite{kazanzides2014open} was fitted with a EndoWrist Suction/Irrigator tool for suctioning.
The cavity and PSM arm were converted into a unified camera frame defined by dVRK's stereo endoscope so that trajectories could be defined in a common reference frame.
To register the cavity, an Aruco marker was attached to it and the pose was solved for using the Aruco library \cite{garrido2014automatic}.
Meanwhile the end-effector of the suction tool was localized in camera frame using our previously developed work, SuPer \cite{li2020super}.

\begin{figure*}[t!]
    \centering
    \vspace{1.8mm}
    \begin{subfigure}{0.195\textwidth}
	\includegraphics[width=1.0\textwidth]{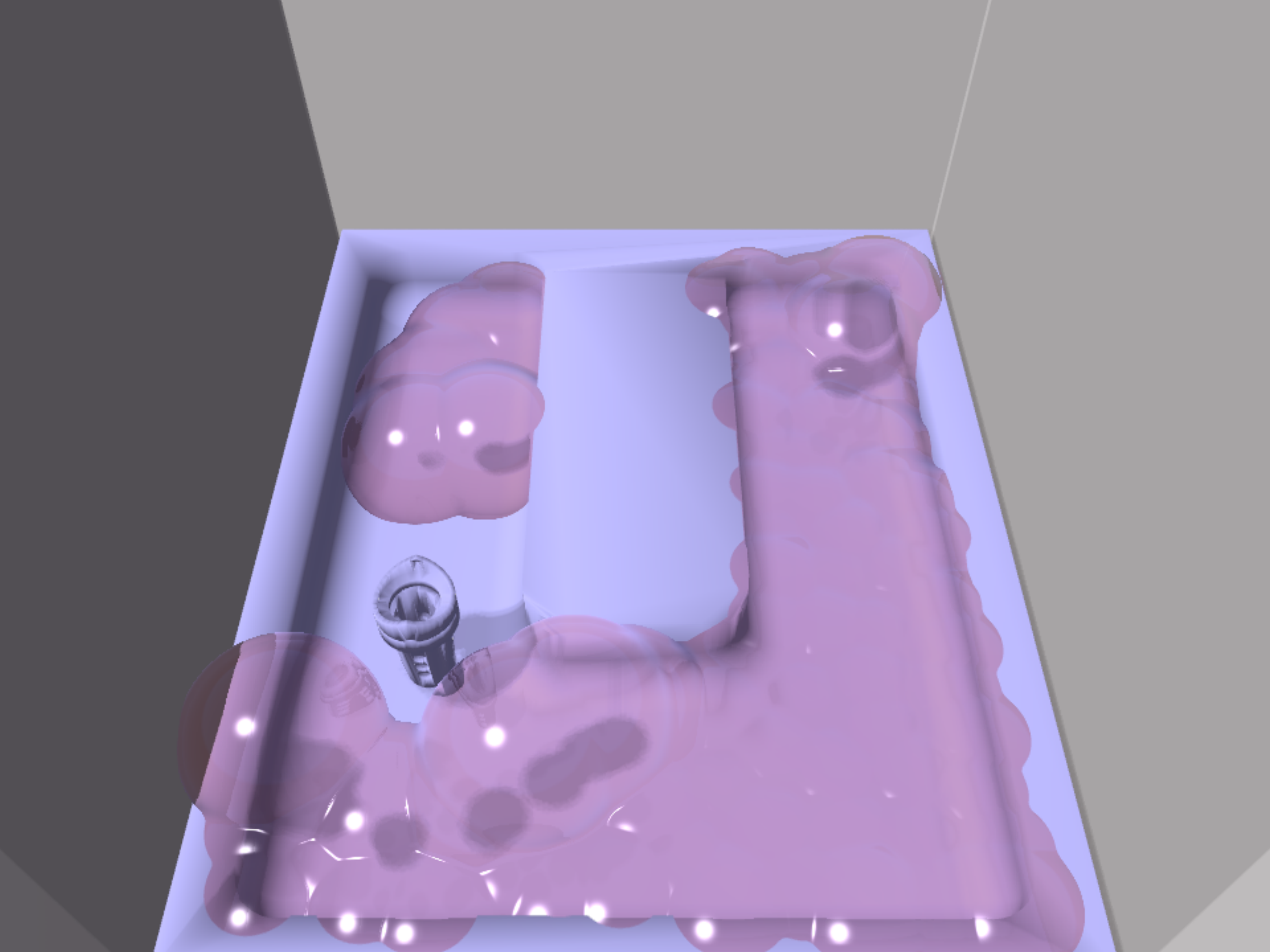}
	\end{subfigure}
    \begin{subfigure}{0.195\textwidth}
	\includegraphics[width=1.0\textwidth]{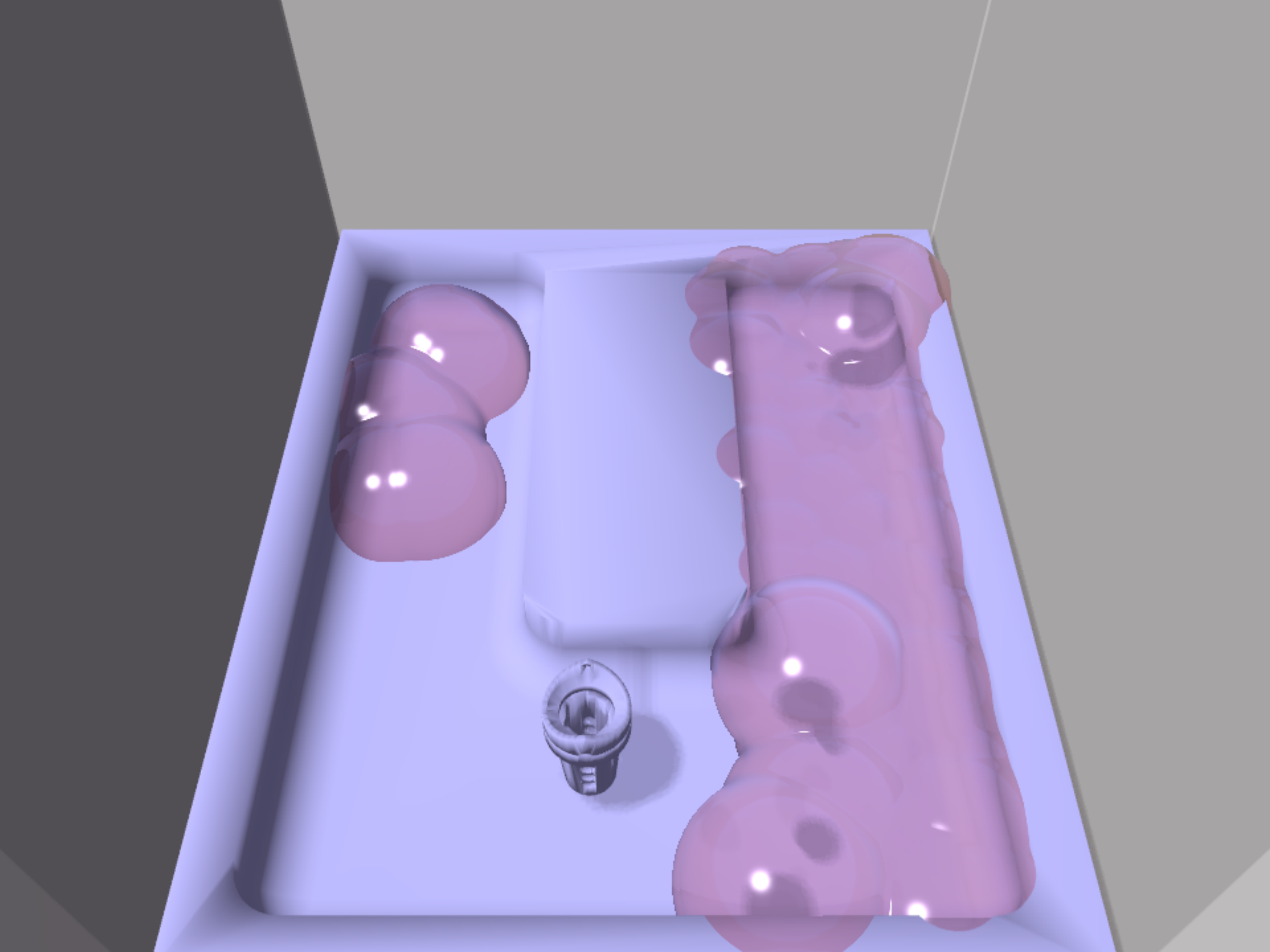}
	\end{subfigure}
	\begin{subfigure}{0.195\textwidth}
	\includegraphics[width=1.0\textwidth]{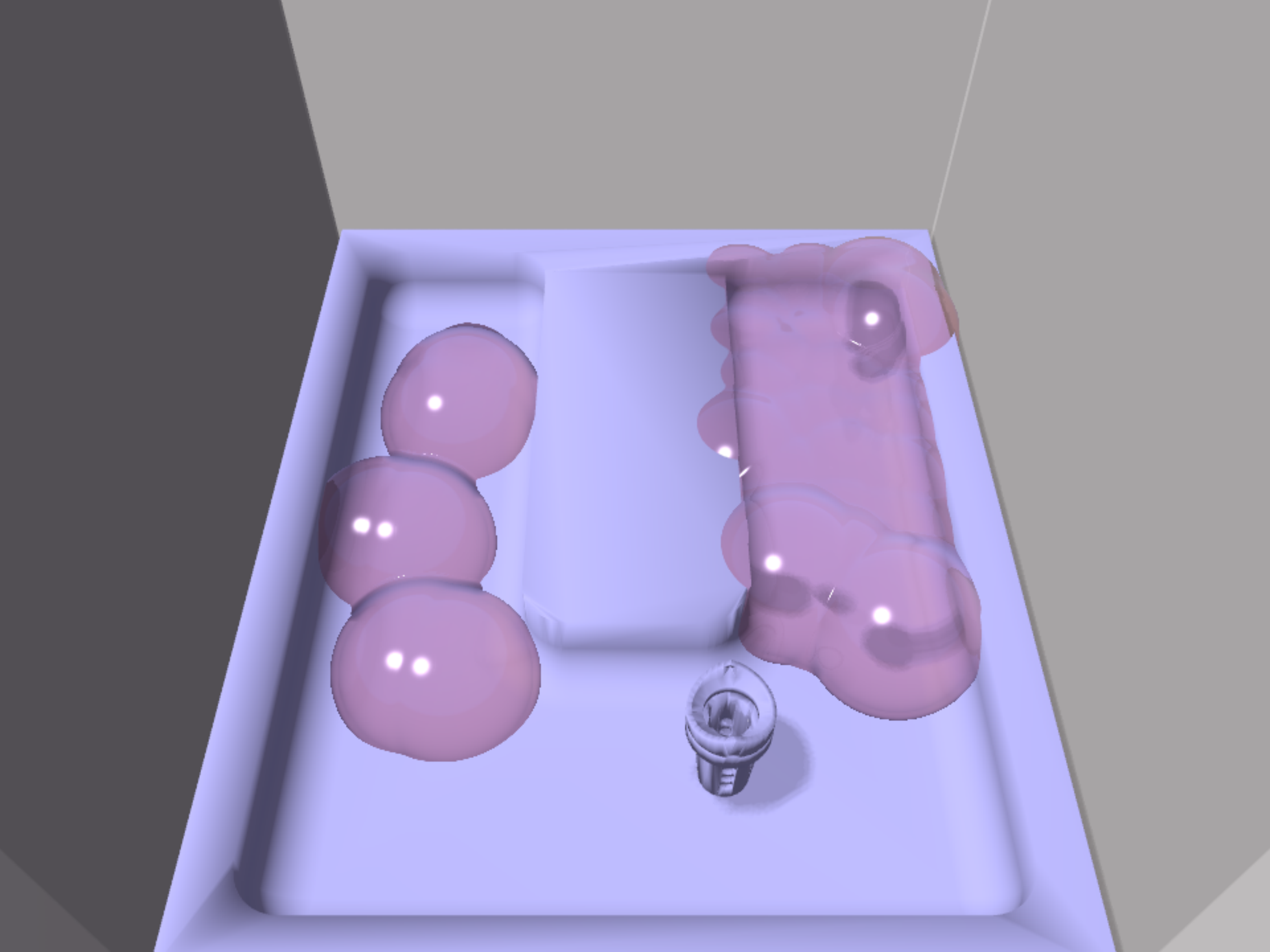}
	\end{subfigure}
	\begin{subfigure}{0.195\textwidth}
	\includegraphics[width=1.0\textwidth]{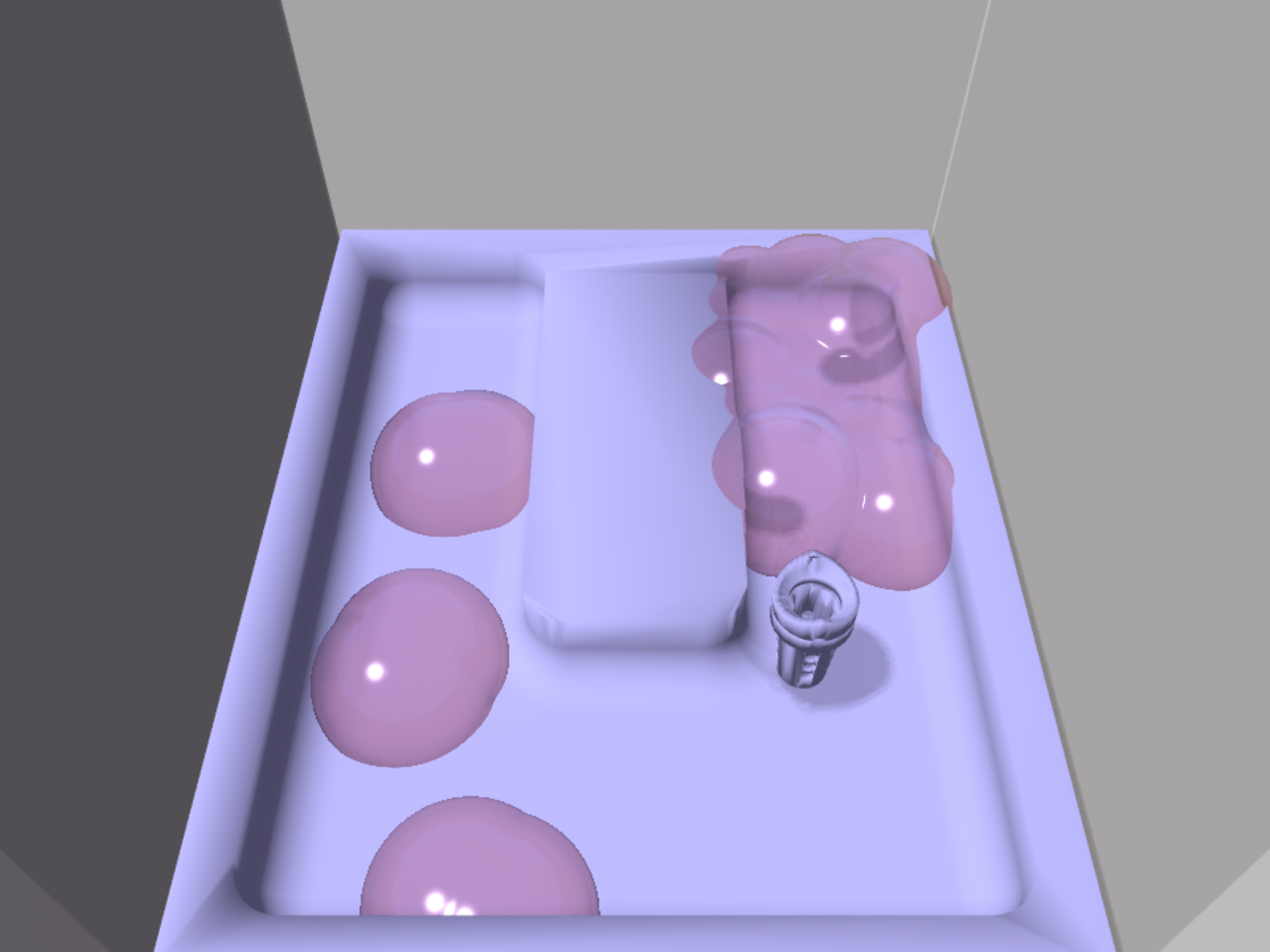}
	\end{subfigure} 
	\begin{subfigure}{0.195\textwidth}
	\includegraphics[width=1.0\textwidth]{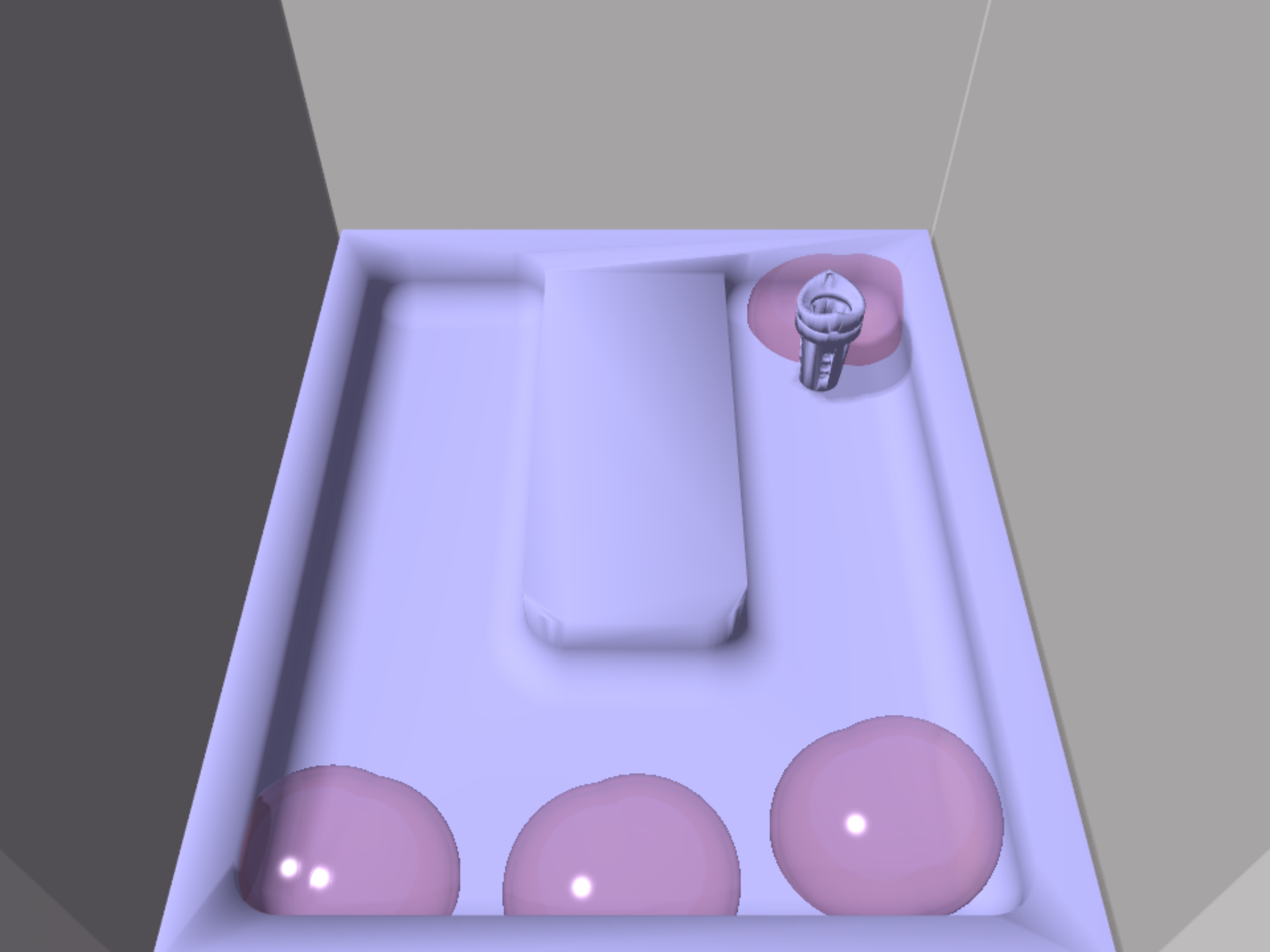}
	\end{subfigure}
	
    \vspace{1mm}
	
	\begin{subfigure}{0.195\textwidth}
	\includegraphics[width=1.0\textwidth]{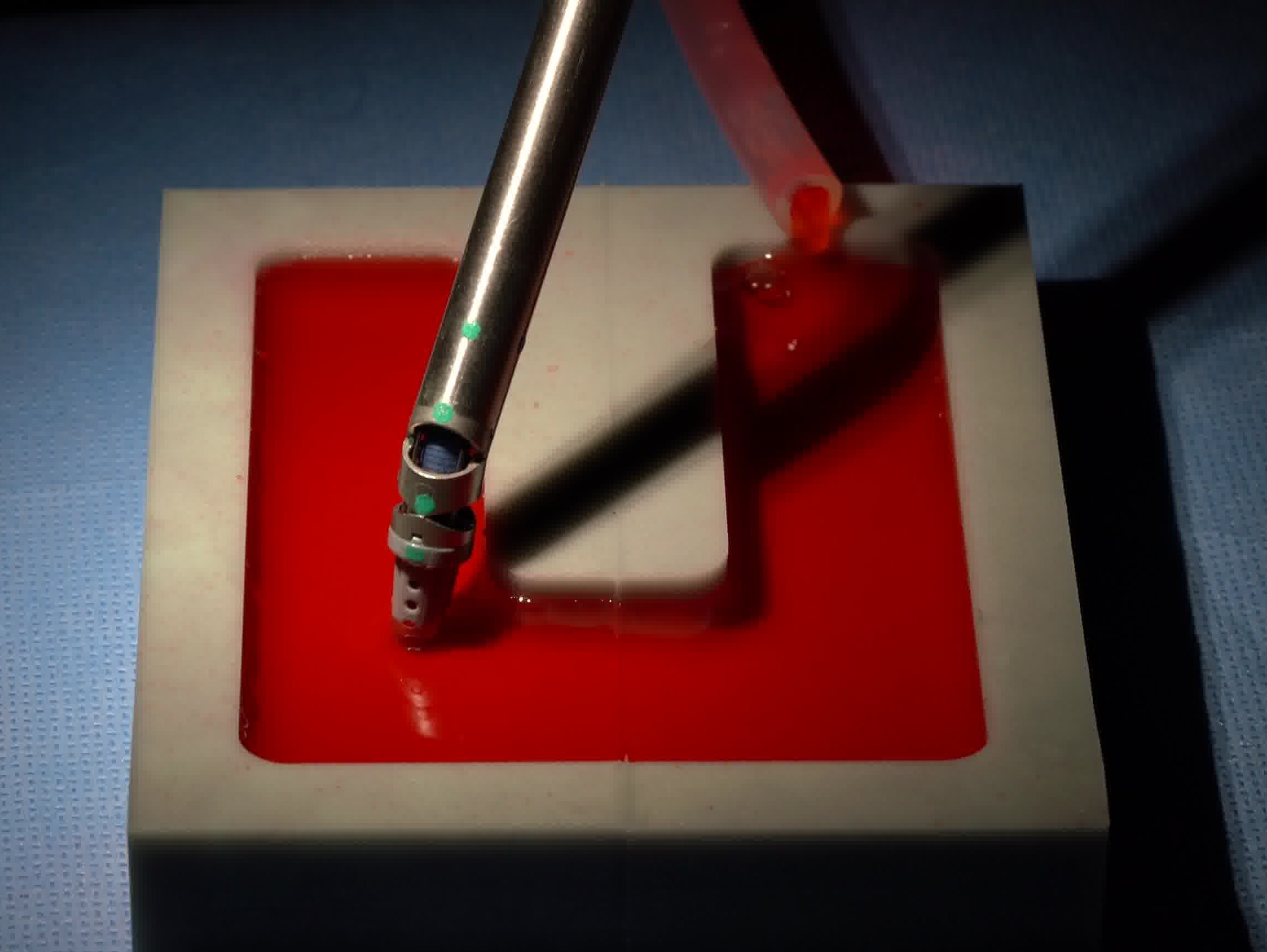}
	\end{subfigure}
    \begin{subfigure}{0.195\textwidth}
	\includegraphics[width=1.0\textwidth]{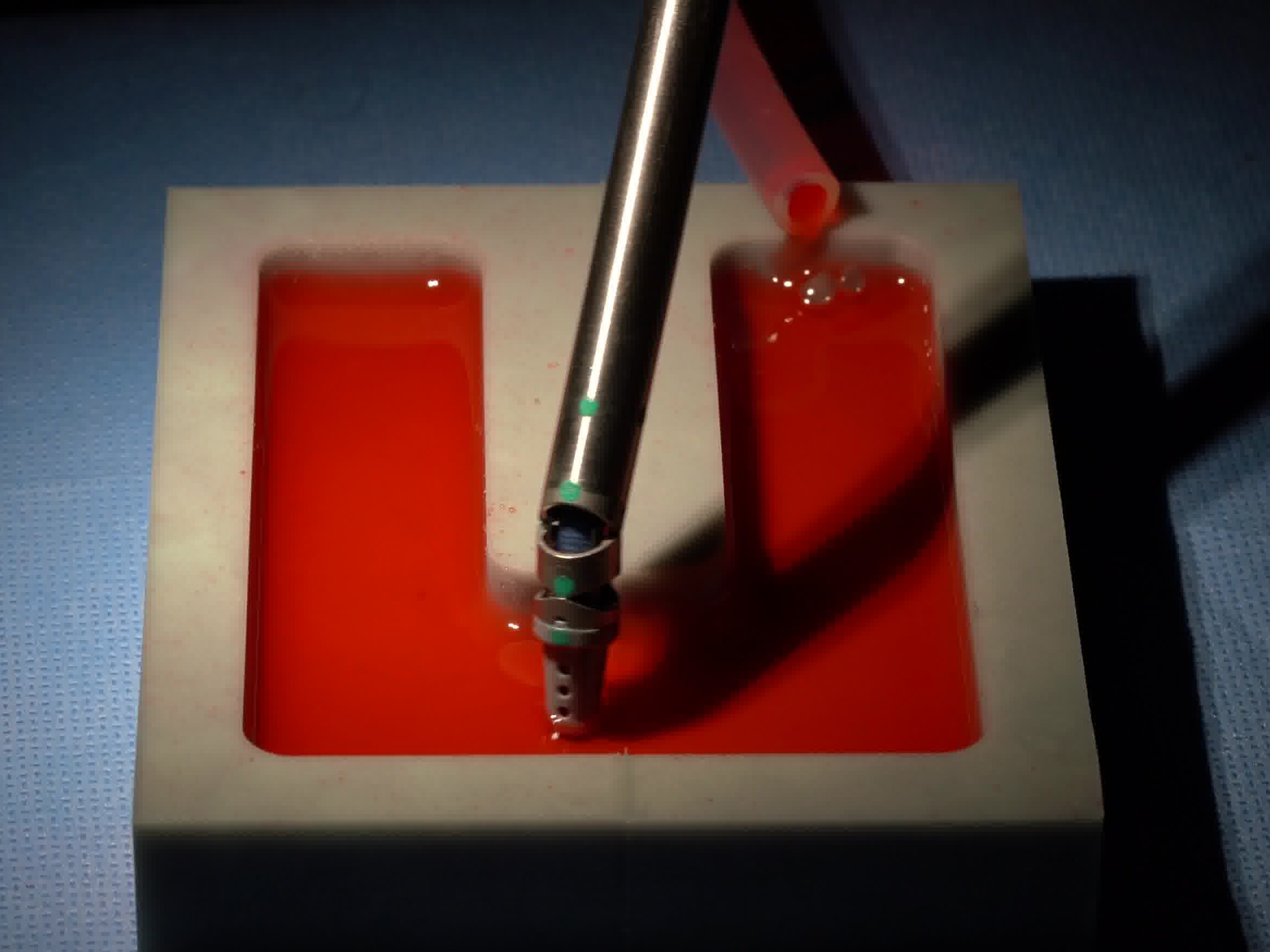}
	\end{subfigure}
	\begin{subfigure}{0.195\textwidth}
	\includegraphics[width=1.0\textwidth]{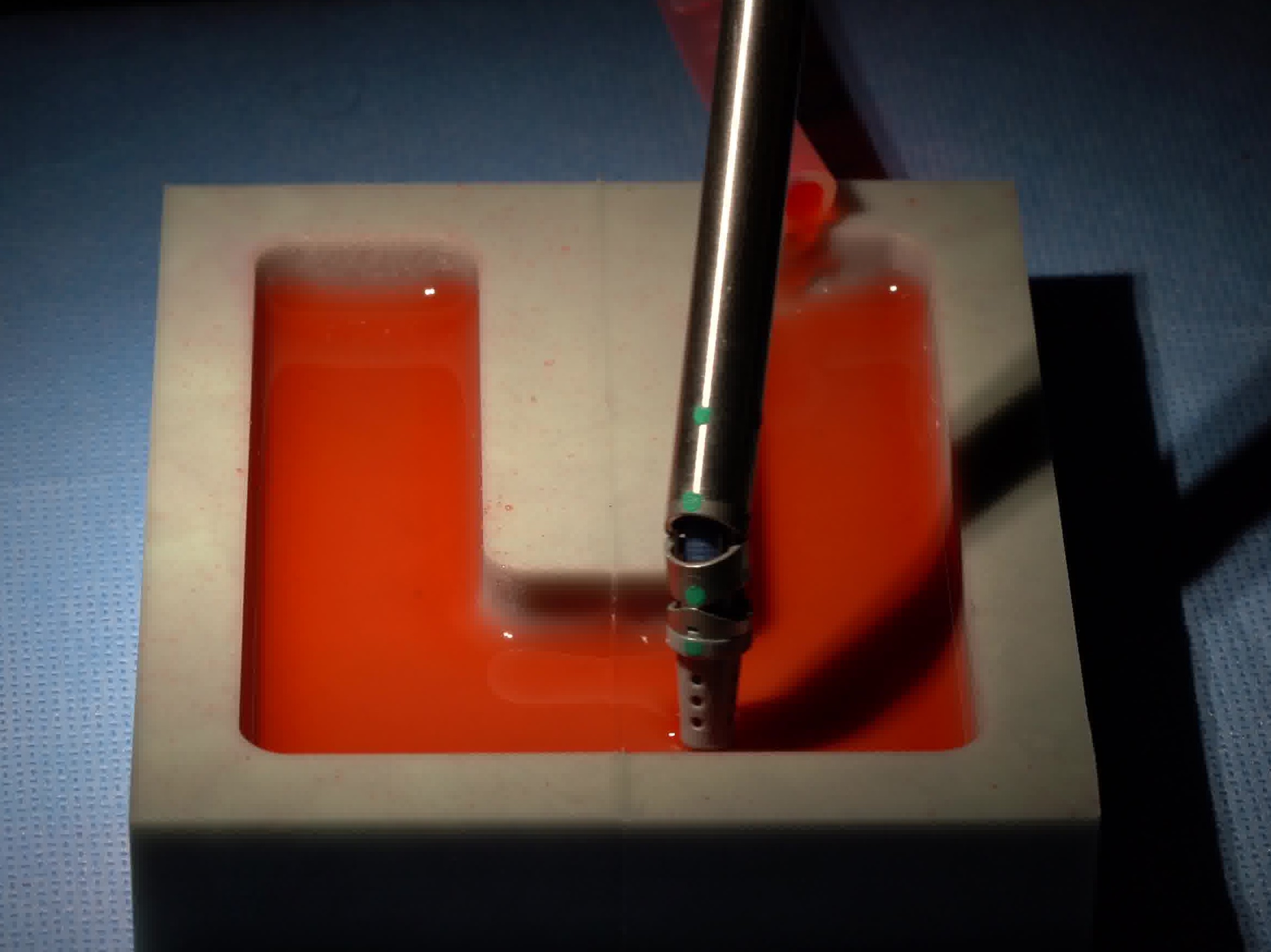}
	\end{subfigure}
	\begin{subfigure}{0.195\textwidth}
	\includegraphics[width=1.0\textwidth]{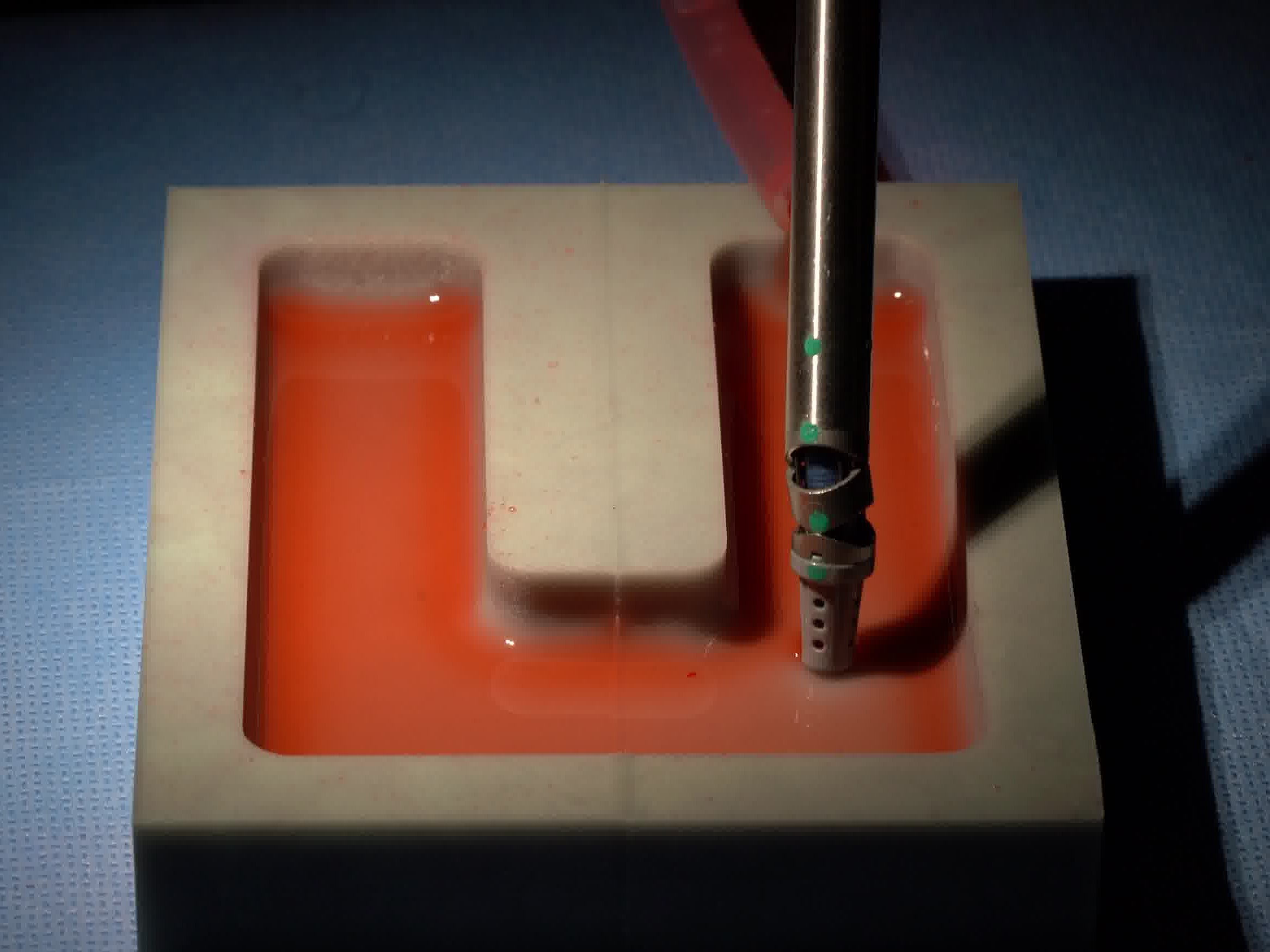}
	\end{subfigure} 
	\begin{subfigure}{0.195\textwidth}
	\includegraphics[width=1.0\textwidth]{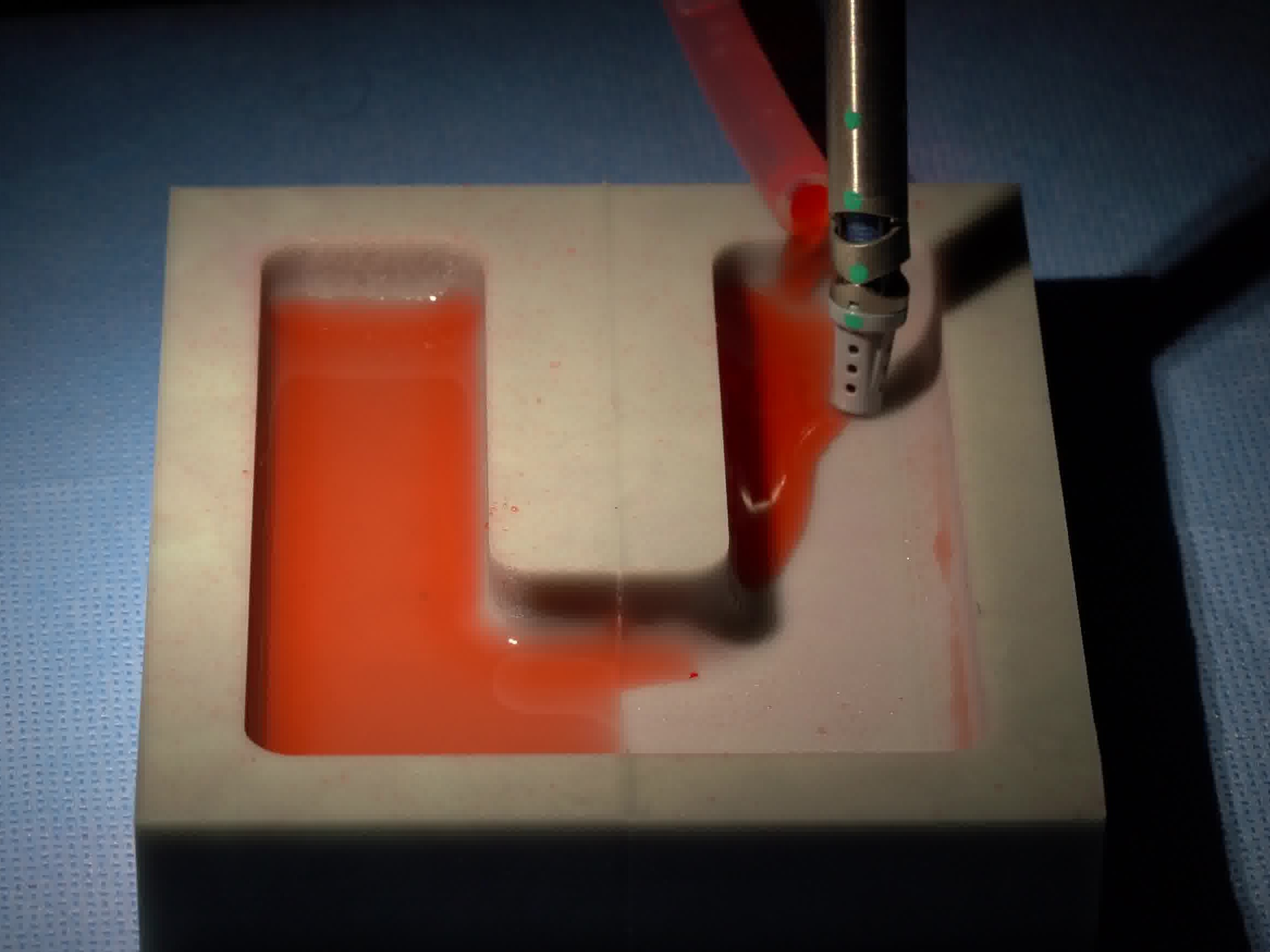}
	\end{subfigure}
	
    \caption{Comparison between the simulated scene (top) and experiments with the real cavity (bottom) for \textit{case 1} using our method.
    Unexpected residual blood occurs in the real world due to unaccounted factors such as surface tension and adhesion.
    Nonetheless, it still effectively removed more than 90\% of the injected blood.
    }
    \label{fig:case1_sim_real_comparison}
    
\end{figure*}

Since trajectories are now required to be described in the common camera frame, the suction tool trajectories, defined in the model's coordinate frame, needs to be properly transformed into the camera frame. This is done by setting the position of the end-effector (suction nozzle) in the PSM base frame, $\mathbf{b}_{t}$, to:
\begin{equation}
    \overline{\mathbf{b}}_{t+1} = \begin{cases} \gamma_s \frac{\mathbf{d}_{t}}{||\mathbf{d}_{t}||} + \overline{\mathbf{b}}_{t}   & \text{if } \mathbf{d}_{t} > \gamma_s\\
    \mathbf{d}_{t} + \overline{\mathbf{b}}_{t} & \text{if } \mathbf{d}_{t} \leq \gamma_s
    \end{cases}
\end{equation}
where $\gamma_s = 0.75$mm is the max step size, the operator $\overline{\cdot} = \begin{bmatrix} \cdot& 1\end{bmatrix}^\top$ gives the homogeneous representation of a point, and the direction, $\mathbf{d}_t$, was computed as
\begin{equation}
    \mathbf{d}_t = \mathbf{T}^b_c \overline{\mathbf{x}}_{e,t} - \overline{\mathbf{b}}_{t}
\end{equation}
with $\mathbf{T}^b_c \in SE(3)$ is the camera to base transform estimated in real-time \cite{li2020super} and $\mathbf{x}_{e,t}$ was the control action defined by the MPC controller.
Inverse kinematics was then used to convert the position, $\mathbf{b}_t$, and orientation to joint angles which were regulated by the robot.

The amount of blood in the cavity was estimated using images from the dVRK's stereo endoscope.
Color segmentation to detect the blood was done by manually setting thresholds in the hue, saturation, value (HSV) color space.
We assume that the concentration of red dye in the water was uniform, which means the attenuation of light through the fluid is proportional to depth by the Beer-Lambert Law \cite{ingle1988spectrochemical}. 
Calibration was performed by taking images of the cavity filled with different volumes of blood and fitting a relationship between the depth of blood and the pixel values in each of the three channels.
The area covered by the blood pixels and the calibrated depth curve in HSV space are used to estimate the volume of blood at each image frame. This model is a simple approximation of the volume of blood, which can only be truly known if one observed the underlying tissue topology before being filled with blood. Naturally, MPC's iterative approach continuously corrects for this volume assumption as more of the tissue topology is revealed, so this first-order approximation is reasonable.

\begin{figure}[t]
    \centering
    \begin{subfigure}{0.45\textwidth}
        \centering
        \includegraphics[width=1.0\textwidth]{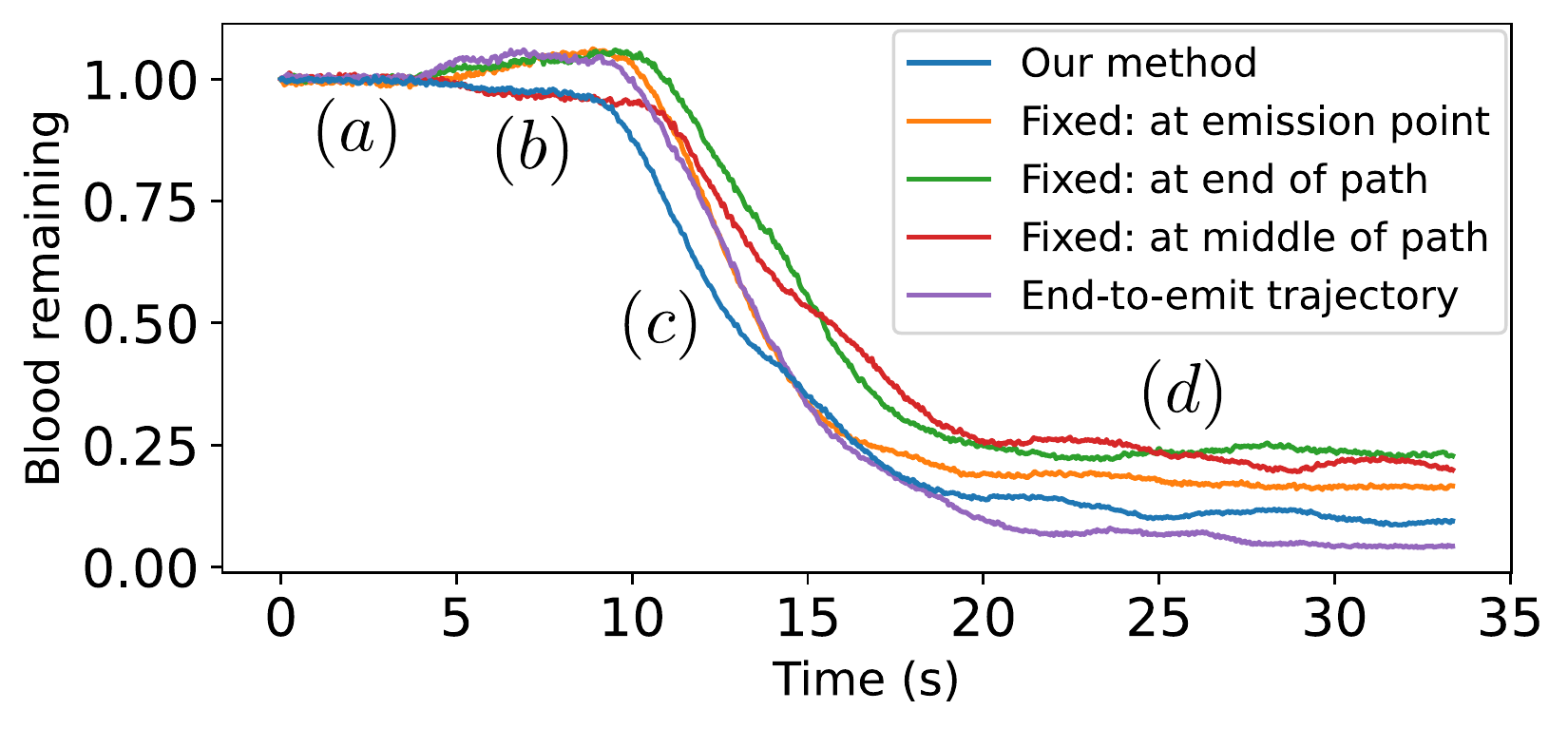}
    \end{subfigure}
    \vspace{-2mm}
    \begin{subfigure}{0.45\textwidth}
        \centering
        \includegraphics[width=1.0\textwidth]{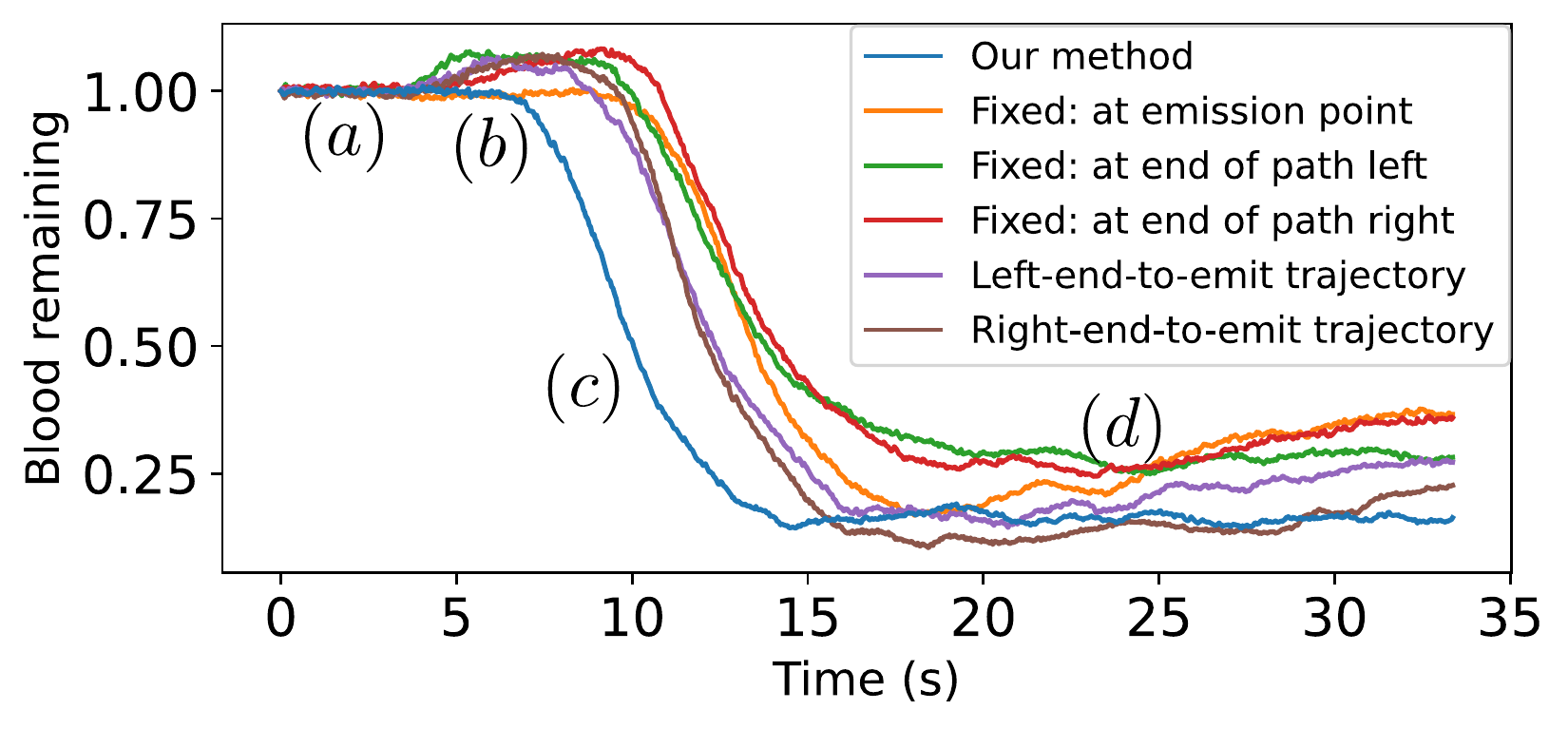}
    \end{subfigure}
    \caption{
    From top to bottom, the plots show suction results from real world experiments for \textit{case 1} and \textit{case 2} respectively, normalized by the initial volume.
    The proposed method reacts more efficiently as seen by the lack of bump at $(b)$, reaches the linear decrease faster at $(c)$, and typically removes the most blood at the end $(d)$.
    }
    \label{fig:real_suction_curve}
\end{figure}

Three trials were repeated per policy and the results are averaged.
In each trial, the cavity was pre-filled with blood, and more blood was continuously injected when suction started to emulate the conditions in the simulated experiments.
The suction curves for experiments with the silicone rubber cavity are shown in Figure \ref{fig:real_suction_curve}. We analyze the suction curves by roughly splitting each trial into 4 distinct stages: $(a)$ where the tool moved to initial position without suctioning $(b)$ when suction and injection began $(c)$ where suction engaged  with the blood for the same time; $(d)$ where suction and injection had been engaged for a long time and the tool reaches the end of the trajectory.


In stage $(b)$, the amount of blood increased as suction started for most of the hand-crafted policies. This means the rate of suction was slower than the rate of blood injection.
Meanwhile the proposed method was able to consistently circumvent this, hence being more efficient.
In stage $(c)$, suction and injection reached an equilibrium where the percentage of remaining blood decreased at a steady rate. 
This rate of decrease was seen to be roughly constant for all policies. 
However, the proposed method was able to reach this state faster due to being more efficient in stage $(b)$.

A discrepancy between the simulation-and-real world was seen at stage $(d)$ where in the real world a thin remaining layer of blood is left over.
An example for \textit{case 1} is shown in Figure \ref{fig:case1_sim_real_comparison}.
In the simulation, the blood particles was able to flow without sticking to the face of the cavity, while a patch of blood was stuck in the left half of the cavity in the real experiment due to surface adhesion.
Nonetheless, our method still results in a low percentage of remaining blood at the end of the task.



%% file: sections/discussion_and_conclusion.tex
In this work we presented a method for incorporating differentiable fluid modeling into autonomous surgical robotics. We applied this method to the surgical sub-task of controlling suction to clear the surgical field of blood during a hemorrhage.
The gradients from the PBF model were used in an MPC framework. The resulting real-world trajectories lead to low percentages of remaining blood despite differences in the fluid parameters between the real and simulated scenes such as surface tension and adhesion of the fluid, and suction strength.
To better overcome these sim-to-real challenges, we plan to integrate visual feedback to initialize and correct the PBF model in real-time using our recently developed blood tracking algorithm \cite{richter2020autonomous}. Finally, our method can be generalized to incorporate different physics models into automating surgical tasks.
Gradients can be derived for not only fluid models, but also for rigid and soft body interactions by viewing their operations as computational graphs in a unified framework \cite{macklin2014unified}.